\documentclass[sigconf]{acmart}
\AtBeginDocument{%
  }

\usepackage{soul}
\usepackage{amsmath,amsfonts}
\usepackage{graphicx}
\usepackage{booktabs}
\usepackage{textcomp}
\usepackage[table,xcdraw]{xcolor}
\usepackage{algorithm, algorithmic}
\usepackage{tabularx}
\usepackage{siunitx}
\usepackage{multirow}
\usepackage{enumitem}
\usepackage{hyperref}
\usepackage{bbding}
\usepackage{float}
\usepackage{balance}
\usepackage{subfig}

\AtBeginDocument{%
  }

\copyrightyear{2026}
\acmYear{2026}
\setcopyright{cc}
\setcctype{by}
\acmConference[KDD '26]{Proceedings of the 32nd ACM SIGKDD Conference on Knowledge Discovery and Data Mining V.2}{August 09--13, 2026}{Jeju Island, Republic of Korea}
\acmBooktitle{Proceedings of the 32nd ACM SIGKDD Conference on Knowledge Discovery and Data Mining V.2 (KDD '26), August 09--13, 2026, Jeju Island, Republic of Korea}
\acmDOI{10.1145/3770855.3818108}
\acmISBN{979-8-4007-2259-2/2026/08}
\copyrightyear{2026}
\acmYear{2026}
\setcopyright{cc}
\setcctype{by-nc-nd}
\acmConference[KDD 2026] {Proceedings of the 32nd ACM SIGKDD Conference on Knowledge Discovery and Data Mining V.2}{August 9--13, 2026}{Jeju Island, Republic of Korea.}
\acmBooktitle{Proceedings of the 32nd ACM SIGKDD Conference on Knowledge Discovery and Data Mining V.2 (KDD 2026), August 9--13, 2026, Jeju Island, Republic of Korea}
\acmISBN{979-8-4007-2259-2/2026/08}
\acmDOI{10.1145/3770855.3818108}

\begin{document}

\title{Bridging Classification and Reconstruction: Cooperative Time Series Anomaly Detection}

\author{Qideng Tang}
\email{tqd18907@nudt.edu.cn}
\additionalaffiliation{%
    \institution{Zhejiang Key Laboratory of Space Information Sensing and Transmission, Hangzhou Dianzi University}
    \city{Hangzhou}
    \country{China}
}
\author{Chaofan Dai}
\email{cfdai@nudt.edu.cn}
\affiliation{%
    \institution{National Key Laboratory of Information Systems Engineering, National University of Defense Technology}
    \city{Changsha}
    \state{Hunan}
    \country{China}
}

\author{Wubin Ma}
\email{wb_ma@nudt.edu.cn}
\author{Yahui Wu}
\email{wuyahui@nudt.edu.cn}
\affiliation{%
    \institution{National Key Laboratory of Information Systems Engineering, National University of Defense Technology}
    \city{Changsha}
    \state{Hunan}
    \country{China}
}

\author{Haohao Zhou}
\email{haohaozhou@nudt.edu.cn}
\authornote{Corresponding authors.}
\affiliation{%
    \institution{National Key Laboratory of Information Systems Engineering, National University of Defense Technology}
    \city{Changsha}
    \state{Hunan}
    \country{China}
}

\author{Tao Zhang}
\email{zhangtao@nudt.edu.cn}
\affiliation{%
    \institution{College of Systems Engineering, National University of Defense Technology}
    \city{Changsha}
    \state{Hunan}
    \country{China}
}

\author{Huan Li}
\email{lihuan.cs@zju.edu.cn}
\affiliation{%
    \institution{Zhejiang University}
    \city{Hangzhou}
    \country{China}
}

\author{Dalin Zhang}
\email{zhangdalin@hdu.edu.cn}
\authornotemark[2]
\affiliation{%
    \institution{Zhejiang Key Laboratory of Space Information Sensing and Transmission, Hangzhou Dianzi University}
    \city{Hangzhou}
    \country{China}
}

\renewcommand{\shortauthors}{Qideng Tang et al.}

\begin{abstract}
    Time series anomaly detection (TSAD) has long been a hot research topic in data mining due to its various applications. Recent studies challenge the effectiveness of popular deep learning methods for TSAD, suggesting their failure in detecting subtle and prolonged anomalies. Outlier Exposure (OE) and Masked Autoencoder (MAE) emerge as two promising paradigms (classification and reconstruction) for solving the above problems. However, OE-based methods are constrained by poor generalization, while MAE-based methods are limited by masking misalignment issues.
    To address these limitations, this paper proposes a novel framework, \textsc{CoAD}, which unifies the two paradigms to leverage their complementary strengths while mitigating their respective weaknesses.
    In this framework, the classification module generates probability-informed soft masks for the reconstruction module, which in turn alleviates the generalization problem of the classification module. This cooperative design enables \textsc{CoAD} to effectively detect subtle and complex anomalies that are often overlooked by existing methods. Additionally, the classification module is carefully designed to resolve issues related to improper classification granularity and the neglect of frequency information.
    Extensive experiments on high-quality benchmark datasets, conducted under rigorous evaluation protocols, demonstrate that \textsc{CoAD} significantly outperforms both state-of-the-art deep learning and traditional data mining methods, highlighting the potential of deep learning in TSAD. Moreover, \textsc{CoAD} is lightweight and substantially faster than existing SOTA methods, demonstrating its practical value for large-scale, real-time applications.
\end{abstract}



\received{20 February 2007}
\received[revised]{12 March 2009}
\received[accepted]{5 June 2009}

\maketitle

\newcommand\kddavailabilityurl{https://doi.org/10.5281/zenodo.20364055}
\ifdefempty{\kddavailabilityurl}{}{
    \begingroup\small\noindent\raggedright\textbf{Resource Availability:}\\
    The source code of this paper has been made publicly available at \url{\kddavailabilityurl} and \url{https://github.com/richard-tang199/CoAD}.
    \endgroup
}

\section{Introduction}

\begin{figure*}[t]
  \centering
  \includegraphics[width=0.8\textwidth]{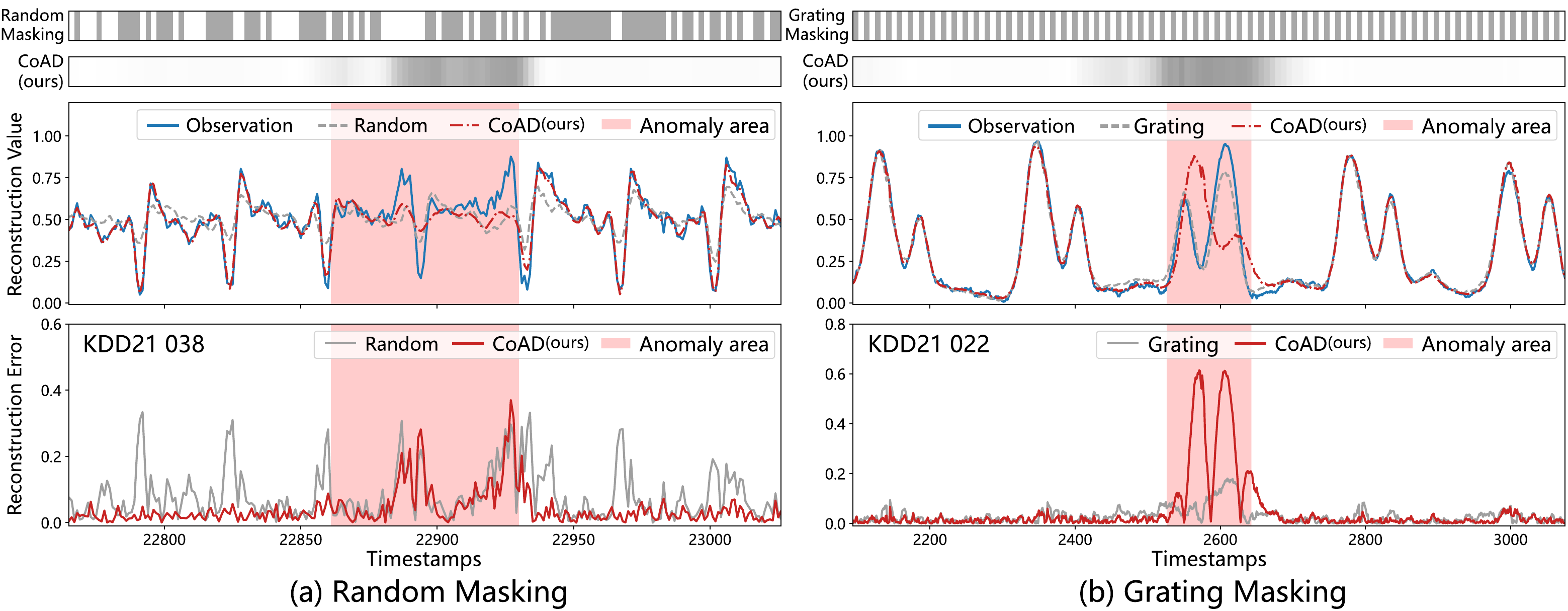}
  \Description{Illustration comparing masking strategies: random masking, grating masking, and the proposed probabilistic soft masking in \textsc{CoAD}, along with the corresponding reconstruction errors.}
  \caption{Comparison between our masking strategy (\textsc{CoAD}) and existing masking strategies. The upper parts visualize the masking.
    Random and grating masking are binarized strategies with shaded areas serving as masks, whereas \textsc{CoAD} is a probabilistic strategy with darker colors indicating stronger masking.}
  \label{fig:insight}
\end{figure*}

Time series anomaly detection aims to identify patterns that deviate from expected behaviors within temporally sequential data and is crucial across numerous applications \cite{blazquezgarciaReviewOutlierAnomaly2022, kimRigorousEvaluationTimeSeries2022,guptaOutlierDetectionTemporal2014,schmidlAnomalyDetectionTime2022}.
In recent years, deep learning has enabled numerous methods for time series anomaly detection (TSAD) \cite{liDeepLearningAnomaly2023, zamanzadehdarbanDeepLearningTime2025, choiDeepLearningAnomaly2021}. However, latest studies \cite{schmidlAnomalyDetectionTime2022,laiRevisitingTimeSeries2021,rewickiItWorthIt2023,liu2024the,Position24,mejriUnsupervisedAnomalyDetection2024,gargEvaluationAnomalyDetection2022} indicate that deep learning-based methods may underperform classical data mining-based methods, especially in detecting subtle and prolonged anomalies \cite{leeExplainableTimeSeries2024, sunUnravelingAnomalyTime2024}. In response, Outlier Exposure (OE) \cite{hendrycks2018deep} and Masked Autoencoders (MAE) \cite{heMaskedAutoencodersAre2022} have emerged as prominent paradigms to solve the above problems\cite{jeongAnomalyBERTSelfSupervisedTransformer2023,wangCutAddPasteTimeSeries2024,xuCalibratedOneclassClassification2024,tang2025MMMA,shentu2025towards,goswami2024moment,fangTemporalFrequencyMaskedAutoencoders2024}. Nevertheless, both paradigms have inherent limitations that can hinder their effectiveness in complex, real-world time series.

\ul{Limitations of OE-based (classification) methods:} \textbf{L1. Heavy reliance on priori knowledge:} OE-based approaches assume the existence of common anomalous patterns and use prior abnormal knowledge to generate pseudo-anomalous samples for classifier training. While effective when real anomalies match the predefined types, this strategy fails to generalize to unseen or unexpected anomalies.
\textbf{L2. Improper classification granularity:} Current OE-based methods operate at either the ``step level'' or the ``window level'', each with drawbacks. Step-level classification \cite{jeongAnomalyBERTSelfSupervisedTransformer2023} assigns anomaly scores to individual time steps by embedding the entire input window in a single forward pass; however, it struggles with longer windows needed for sufficient context \cite{tang2025MMMA}. In contrast, window-level classification \cite{wangCutAddPasteTimeSeries2024, xuCalibratedOneclassClassification2024} predicts anomalies for the whole window and slides it to generate stepwise scores. Although this captures longer contexts, subtle anomalies can be obscured by predominantly normal patterns, leading to missed detections.
\textbf{L3. Neglect of frequency-domain information:} Most OE-based methods operate solely in the time domain, ignoring the frequency domain where some anomalies may be more pronounced. Consequently, frequency-sensitive anomalies that are subtle or ambiguous in the time domain may go undetected.

\ul{Limitations of MAE-based (reconstruction) methods:} \textbf{L4. Masking misalignment with anomaly locations:} MAE-based methods model normal patterns by reconstructing masked patches from unmasked ones and assign anomaly scores based on reconstruction errors. Ideally, the masking should target potentially anomalous regions while leaving normal patches unmasked, allowing the model to rely on surrounding normal patterns for reconstruction. However, as shown in Figure \ref{fig:insight}, existing methods typically use random masking \cite{tang2025MMMA, shentu2025towards} or grating masking \cite{chenImDiffusionImputedDiffusion2023}, without considering patch semantics, and indiscriminately mask normal and anomalous regions. Consequently, the model may reproduce anomalous patterns, leading to false alarms in normal regions \cite{tang2025MMMA} (large reconstruction errors in Figure \ref{fig:insight}(a)) or miss anomalies by accurately reconstructing anomalous regions \cite{yaoOneforAllProposalMasked2023} (Figure \ref{fig:insight}(b)).

To overcome these limitations, we propose \textsc{CoAD}, a cooperative anomaly detection framework that unifies the strengths of classification- and reconstruction-based methods. At the core of \textsc{CoAD} is a guided soft masking mechanism, which leverages OE-based classification to inform the masking for MAE-based reconstruction. Unlike conventional random or uniform masking, \textsc{CoAD} applies probability-informed soft masking, where all patches are masked, but those more likely to be anomalous are masked more heavily (see Figure~\ref{fig:insight}). This suppresses anomaly-related cues during reconstruction, yielding more accurate anomaly scores and effectively \textbf{\textit{addressing L4}}.
Since MAE-based reconstruction can restore normal patterns within anomalous regions, it in turn provides a ``quasi-normal reference'' for OE-based classification. This insight motivates the design of a residual-based classification module that leverages the discrepancy between original patches and their reconstructed counterparts in the feature space as generalizable representations for anomaly discrimination, thereby facilitating the detection of novel anomalies. Furthermore, the MAE-based reconstruction is constrained by OE to learn only the normal data distribution and therefore fails to reproduce unseen anomalies. During inference, the classification and reconstruction modules operate collaboratively to detect anomalies. The overall cooperative mechanism enables \textsc{CoAD} to identify previously unseen anomalies, thereby \textbf{\textit{addressing L1}}.
To support this cooperative strategy, \textsc{CoAD} introduces a patch-level, dual-branch time-frequency classification module. The input sequence is divided into non-overlapping patches, and each patch is classified based on features extracted from both the time and frequency domains. This design offers two advantages: it captures both long- and short-term contextual information via intra- and inter-patch correlations (\textbf{\textit{addressing L2}}), and it incorporates frequency-domain patterns often overlooked by existing methods (\textbf{\textit{addressing L3}}).

In light of recent concerns~\cite{liu2024the, wuCurrentTimeSeries2021, eamonnKeoghIrrational21, 2024Fundamental} regarding the reliability of experiments in many existing studies, we conduct a rigorous evaluation using the highest-quality datasets and the most robust metrics available. Experimental results show that \textsc{CoAD} consistently outperforms \textbf{24 SOTA methods} across \textbf{314 datasets}, achieving significantly superior performance. Moreover, qualitative analyses confirm that \textsc{CoAD} can effectively detect subtle and challenging anomalies that are often overlooked by existing methods, including cases that are difficult even for human experts. Furthermore, \textsc{CoAD} runs orders of magnitude faster than existing methods, highlighting its remarkable efficiency and scalability.

\section{Related Work}\label{sec:related}
\subsection{Data Mining vs Deep Learning}
Recent studies \cite{rewickiItWorthIt2023, liu2024the, mejriUnsupervisedAnomalyDetection2024} challenge the effectiveness of popular deep learning-based methods for TSAD, suggesting that classical data mining methods such as discord-based methods (e.g., Matrix Profile \cite{nakamuraMERLINParameterFreeDiscovery2020} and DAMP \cite{luDAMPAccurateTime2023}) and clustering-based methods (e.g., SAND \cite{boniolSANDStreamingSubsequence2021} and KShapeAD \cite{liu2024the}), outperform deep learning methods. These studies \cite{liu2024the, sunUnravelingAnomalyTime2024,tang2025MMMA} particularly highlight that deep learning methods fail to detect subtle and prolonged anomalies due to their inability to learn a discerning boundary between normal and subtle abnormal samples \cite{wangCutAddPasteTimeSeries2024,xuCalibratedOneclassClassification2024}.

\subsection{OE-based Methods}
OE-based methods assume that common anomalous patterns exist in time series and aim to train a binary classifier based on generated pseudo-anomalies. By explicitly injecting prior abnormal knowledge into the model, these methods enable the learning of a clearer decision boundary between normal and abnormal samples.
AnomalyBERT \cite{jeongAnomalyBERTSelfSupervisedTransformer2023} is a pioneering work that assumes four representative types of anomalies in time series and employs a BERT-like structure to extract features and perform anomaly classification. Although it achieves promising performance, it suffers from heavy reliance on pre-assumed anomalous patterns and often generalizes poorly to unseen anomaly types.
Subsequent research attempts to address this issue by either replacing subsequences with random segments clipped from other positions as pre-assumed anomalies \cite{wangCutAddPasteTimeSeries2024}, or by employing a one-class classifier to additionally measure the distance to normal patterns \cite{xuCalibratedOneclassClassification2024}. Although these techniques appear promising, their advances do not lead to substantial improvements (see Table~\ref{tab:main_result}). Generalization still remains the primary challenge for the OE-based methods.

\subsection{MAE-based Methods}
MAE-based methods capture the normal temporal dependencies in time series by learning to reconstruct masked parts from unmasked parts. These methods \cite{tang2025MMMA, shentu2025towards, goswami2024moment, fangTemporalFrequencyMaskedAutoencoders2024} demonstrate strong capability in modeling long sequences and achieve significant performance improvements compared to traditional reconstruction-based methods.
The masking strategy is the core technical component of MAE-based methods. The most intuitive and widely adopted approach is random masking \cite{tang2025MMMA, shentu2025towards, chenImDiffusionImputedDiffusion2023}. However, purely random masking can excessively obscure local normal information, making it difficult for the model to reconstruct normal regions accurately. An alternative is grating masking \cite{chenImDiffusionImputedDiffusion2023}, which imposes a regular masking pattern. Nevertheless, this approach may result in substantial leakage of anomalous information, making the model to generate good reconstruction for abnormal patches, causing failure of anomaly detection. In the ideal case, we expect the suspicious anomalies are masked as much as possible.
TFMAE \cite{fangTemporalFrequencyMaskedAutoencoders2024} takes a preliminary step in this direction by masking subsequences with high variance within the input window; however, anomalies do not always correspond to high-variance areas.
\section{Methodology}\label{sec:method}

\subsection{Task Description and Notations}
Let $\mathbf{S} = \{x_1, x_2, \ldots, x_\mathtt{L} \}$ denote a time series of length $\mathtt{L}$, where $x_i$ represents the observation at time step $i$. The goal of time series anomaly detection is to assign an anomaly score $A(x_i) \in \mathbb{R}$ to each observation $x_i \in \mathbf{S}$, where a higher value of $A(x_i)$ indicates a greater likelihood that $x_i$ is anomalous.
Following conventions \cite{liDeepLearningAnomaly2023, zamanzadehdarbanDeepLearningTime2025, mejriUnsupervisedAnomalyDetection2024}, we apply a sliding window to segment the raw time series $\mathbf{S}$ into a collection of windows, $\{\mathbf{X}_1, \mathbf{X}_2, \ldots, \mathbf{X}_\mathtt{K} \}$, where each window $\mathbf{X}_n = \{x_{n,1}, x_{n,2}, \ldots, x_{n,\mathtt{T}}\}$ consists of $\mathtt{T}$ consecutive time steps and serves as a model input. For simplicity, we omit the window index $n$ and denote a generic input window as $\mathbf{X} = \{x_1, x_2, \ldots, x_\mathtt{T}\}$.

\begin{figure*}[!t]
  \centering
  \includegraphics[width=0.9\textwidth]{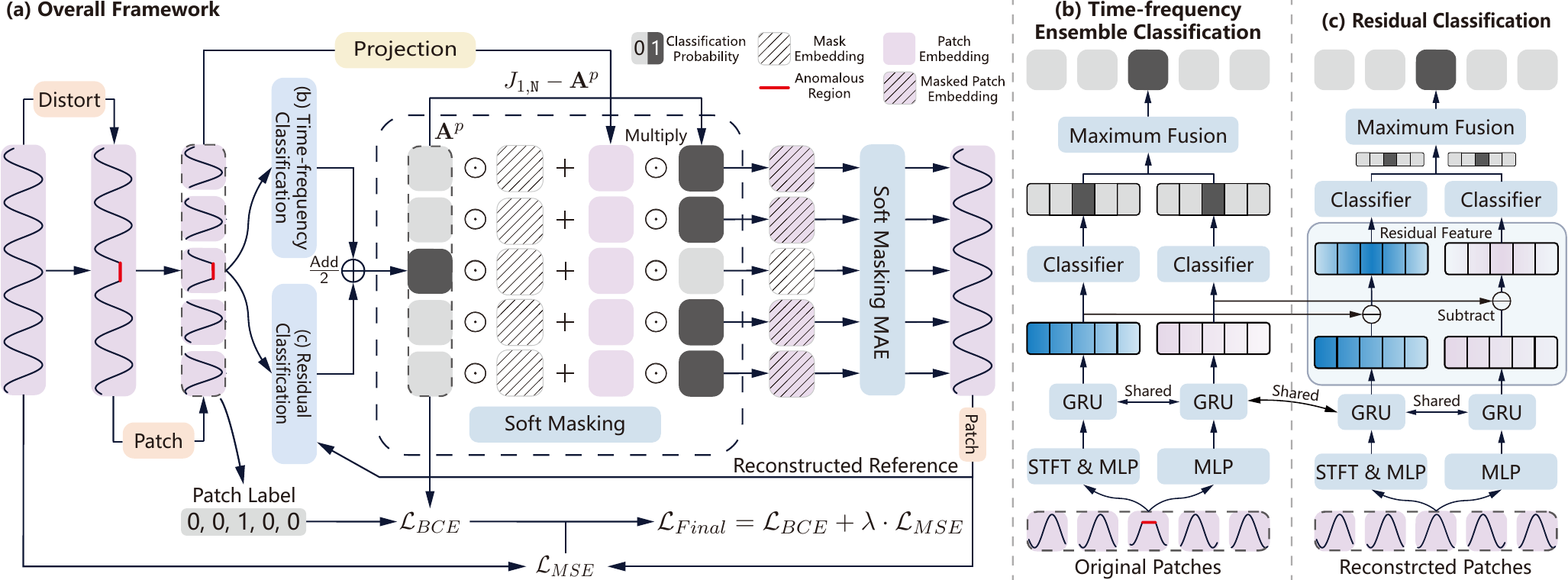}
  \caption{The \textsc{CoAD} framework. (a) Overall framework. (b) Time-Frequency ensemble classification. (c) Residual classification.}
  \label{fig:framework}
\end{figure*}

\subsection{Cooperative Anomaly Detection  Framework}\label{sec:framework}

We propose \textsc{CoAD}, a cooperative framework that seamlessly integrates classification and reconstruction to leverage their complementary strengths and overcome their individual limitations. As illustrated in Figure~\ref{fig:framework}(a), \textsc{CoAD} is built on the insight that the classification module guides the reconstruction process to enhance detection accuracy, while the reconstruction module provides normal references that facilitate the identification of unseen anomalies.
Specifically, the framework employs an anomaly classifier trained on predefined anomaly patterns to produce patch-wise anomaly probabilities. These probabilities are then used to guide the masking process of the MAE module, enabling the maximal suppression of anomalous information (Section~\ref{sec:rec}).
The reconstructed patches provided by the MAE module serve as references to extract generalizable residual features for a second stage of anomaly classification (Section~\ref{sec:residual}).
All components are jointly trained with a weighted loss combining Binary Cross Entropy (for classification) and Mean Squared Error (for reconstruction), ensuring mutual optimization and synergy (Section~\ref{sec:train}).

To further enhance performance, the classification module incorporates two key components. One is patch-level classification, which is made feasible by the cooperative framework. Since the reconstruction module generates fine-grained anomaly scores at the timestamp level, the classification module can operate on non-overlapping patches. This substantially reduces the input sequence length, reducing computational burden and enabling the model to more effectively capture intra- and inter-patch contextual dependencies to find anomalies (see Section~\ref{ssec:patch}). The other is a time-frequency dual-branch ensemble that aggregates complementary temporal and frequency features, boosting the model's capacity to detect complex and subtle anomalies (see Section~\ref{ssec:tfe}).
\subsection{Mask Generation via Patch-level Time-frequency Classification}\label{sec:cls}
\subsubsection{Distortion and Patching}\label{ssec:patch}
The classification module is trained on synthetic anomalies to incorporate generalized prior knowledge regarding anomalous patterns. Specifically, the input window is stochastically distorted to generate simulated anomalies and is then partitioned into non-overlapping patches.

\textbf{Distortion:} Following established methodologies ~\cite{jeongAnomalyBERTSelfSupervisedTransformer2023,shentu2025towards}, we simulate four types of time series anomalies: Uniform Replacement, Mirror Flip, Length Scale, and Jittering. These distortion methods are inspired by the most prevalent types of anomalies~\cite {blazquezgarciaReviewOutlierAnomaly2022,mejriUnsupervisedAnomalyDetection2024,zamanzadehdarbanDeepLearningTime2025}. Formally, we randomly select a segment of the input window $\mathbf{X}$, with a length ranging from 0 to one dominant period, and inject one of the four types of anomalies, yielding a distorted series denoted as $\mathbf{\tilde{X}}$. \textbf{Details of these distortion strategies are available in Appendix~\ref{sec:simulated_anomalies}}.

\begin{figure}[t]
  \centering
  \includegraphics[width=\columnwidth]{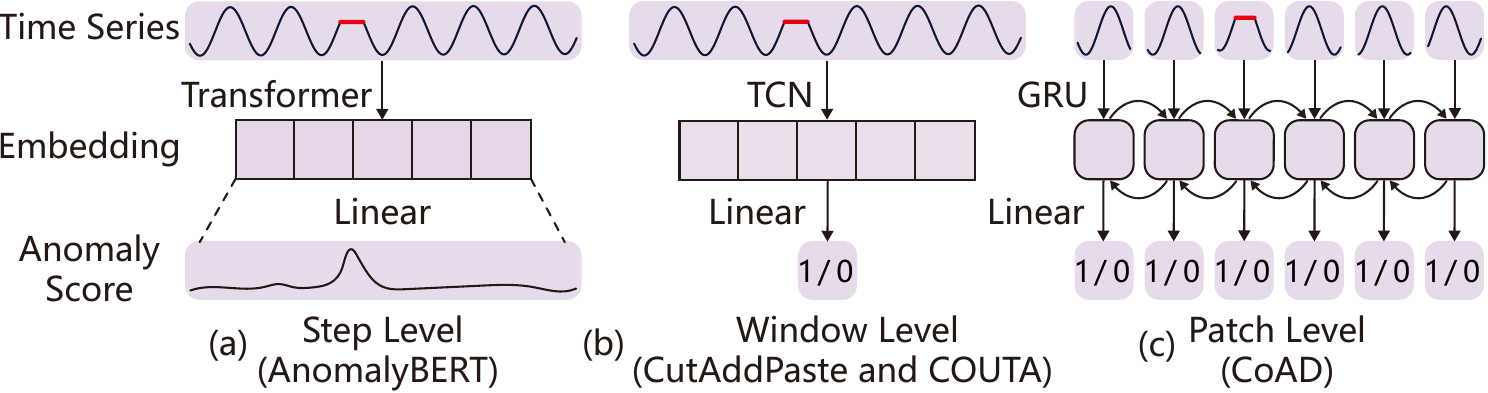}
  \caption{Different levels of classification granularity.}
  \label{fig:classify_example}
  \vspace{-0.5cm}
\end{figure}

\textbf{Patching:} As illustrated in Figure~\ref{fig:classify_example}(a,b), existing classification-based methods typically adopt either ``step-level'' or ``window-level'' classification granularity, both of which have inherent limitations.
Step-level methods encode the entire input window into a single latent representation, then use a decoder to produce anomaly scores for each time step~\cite{jeongAnomalyBERTSelfSupervisedTransformer2023}. However, producing accurate fine-grained scores from a single embedding is especially difficult for long sequences, which are necessary to capture sufficient contextual information~\cite{tang2025MMMA,nie2023a}.
Window-level methods also use a single embedding for the entire input window but output only a binary label indicating whether any anomaly exists in the window~\cite{wangCutAddPasteTimeSeries2024,xuCalibratedOneclassClassification2024}. This coarse prediction can easily overlook short-duration or subtle anomalies that are masked by dominant normal patterns.

To address the above issues, we adopt a ``patch-level'' classification strategy. Specifically, an input window is segmented into non-overlapping patches $ \mathbf{\tilde{X}}^p= [\tilde{x}^p_1,\tilde{x}^p_2,\ldots,\tilde{x}^p_\mathtt{N}] \in \mathbb{R}^{\mathtt{P}\times \mathtt{N}}$, where $\mathtt{N}$ represents the number of patches and $\tilde{x}^p_i$ denotes a patch containing $\mathtt{P}$ time points.
As shown in Figure~\ref{fig:classify_example}(c), each patch is first embedded via a linear layer to model short-term dependencies, and then all embeddings are interacted using a Bi-GRU module to extract long-term correlations, followed by a shared linear layer that classifies whether a patch contains anomalies (detailed in Section \ref{ssec:tfe}).
This patch-level design strikes a balance between the granularity of step-level and window-level methods. Crucially, it is made feasible by our cooperative framework: since the reconstruction module provides fine-grained anomaly scores at the step level, the classification module does not need to produce scores for each individual time step. This decoupling enables non-overlapping patches, which substantially reduces the sequence length and enables the model to more effectively learn both intra- and inter-patch context with reduced computational overhead.

\begin{figure*}[t]
  \centering
  \includegraphics[width=\textwidth]{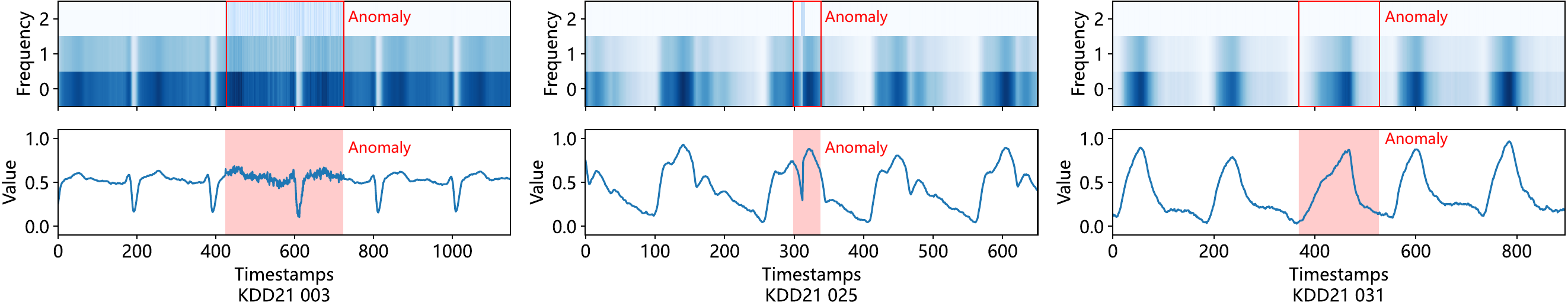}
  \caption{Comparison of frequency domain features between normal and anomalous regions. The upper panel displays the STFT spectrogram, with darker colors indicating higher amplitudes. Anomalies that are difficult to detect in the time domain exhibit more distinctive and discriminative patterns in the frequency domain.}
  \label{fig:stft}
  \vspace{-0.5cm}
\end{figure*}

\subsubsection{Time-frequency Ensemble Classification}\label{ssec:tfe}
Time-frequency analysis is widely used in time series research. As illustrated in Figure~\ref{fig:stft}, certain anomalies that are difficult to distinguish in the time domain exhibit more salient patterns in the frequency domain. While existing classification-based anomaly detection methods generally overlook frequency features, some reconstruction-based approaches have incorporated time-frequency representations to improve detection performance \cite{fangTemporalFrequencyMaskedAutoencoders2024, wu2024catch, wangRevisitingVAEUnsupervised2024a, zhangTFADDecompositionTime2022}.

However, these reconstruction-based approaches face two limitations.
\textit{i)} As observed in Figure~\ref{fig:stft} and corroborated by prior studies \cite{wu2024catch, zhangTFADDecompositionTime2022}, frequency amplitudes vary considerably across different bands, typically higher in low-frequency and lower in high-frequency regions. This uneven distribution complicates accurate reconstruction across all frequency bands, often resulting in large relative errors in high-frequency components. Consequently, normal high-frequency patterns may be misidentified as anomalies due to large reconstruction errors.
To mitigate this issue, we propose using frequency-domain features for classification rather than as reconstruction targets.
\textit{ii)} Most existing works \cite{wangRevisitingVAEUnsupervised2024a, zhangTFADDecompositionTime2022, wu2024catch} extract frequency features via the Fast Fourier Transform (FFT), which provides fine-grained frequency resolution but only coarse (window-level) resolution in the time domain. To address this, we adopt the Short-Time Fourier Transform (STFT), which enables finer temporal localization of frequency patterns. By integrating both time- and frequency-domain representations, our dual-branch classification module leverages complementary information to enhance anomaly detection capability. The structure of the time-frequency ensemble classification is demonstrated in Figure \ref{fig:framework}(b), and the details are described as follows:

\textbf{Frequency branch:}
In the frequency classification branch, we apply the STFT to the entire input window $\mathbf{\tilde{X}}$ to obtain the frequency-domain representation $\mathbf{\tilde{X}}_f \in \mathbb{R}^{2\mathtt{K} \times \mathtt{T}}$, where $\mathtt{K}$ denotes the number of frequency bins. The real and imaginary parts of the complex-valued STFT output are concatenated directly. We then segment $\mathbf{\tilde{X}}_f$ into non-overlapping patches $\mathbf{\tilde{X}}_f^p = [\tilde{x}^p_{f,1},\tilde{x}^p_{f,2},\ldots,\tilde{x}^p_{f,\mathtt{N}}]\in \mathbb{R}^{2\mathtt{K}\mathtt{P} \times \mathtt{N}}$, where $\mathtt{P}$ is the patch length and $\mathtt{N}$ is the number of patches.

To model inter-patch dependencies, each frequency patch is first linearly projected into a hidden space and then processed by a GRU encoder:
\begin{equation}
  \mathbf{\tilde{H}}_f^{p} = \operatorname{GRU}\left( \mathbf{W}_f^p \mathbf{\tilde{X}}_f^p \right),
\end{equation}
where $\mathbf{W}_f^p \in \mathbb{R}^{\mathtt{H} \times 2\mathtt{K}\mathtt{P}}$ is a learnable projection matrix and $\mathtt{H}$ is the hidden dimension.
The output $\mathbf{\tilde{H}}_f^{p}$ is then passed through a linear classifier layer followed by a sigmoid activation to generate the anomaly probability for each patch:
\begin{equation}
  \mathbf{A}_f^p = \sigma\left(\mathbf{W}_{\mathbf{f}} \mathbf{\tilde{H}}_f^{p}\right),
\end{equation}
where $\mathbf{W}_{\mathbf{f}} \in \mathbb{R}^{1 \times \mathtt{H}}$ is a learnable weight vector, $\sigma (\cdot)$ represents the sigmoid activation function, and $\mathbf{A}_f^p = [ a^p_{f,1}, a^p_{f,2}, \ldots,a^p_{f,\mathtt{N}}]$ denotes the predicted anomaly probabilities for the frequency domain patches $\mathbf{\tilde{X}}_f^p$.

\textbf{Time branch:} The patch set $\tilde{\mathbf{X}}^{p}$ is directly projected via a linear layer and then fed into a GRU encoder. The obtained features $\mathbf{\tilde{H}}_t^{p}$ are subsequently processed through a linear layer followed by a sigmoid activation to produce patch-wise anomaly probabilities:
\begin{equation}
  \mathbf{\tilde{H}}_t^{p} = \operatorname{GRU}\left( \mathbf{W}_t^p \mathbf{\tilde{X}}^p \right), \, \,
  \mathbf{A}_t^p = \sigma\left(\mathbf{W}_{\mathbf{t}}\mathbf{\tilde{H}}_t^{p} \right),
\end{equation}
where $\mathbf{A}_t^p=[a_{t,1}^p, a_{t,2}^p, \ldots, a_{t,\mathtt{N}}^p]$ denotes the predicted anomaly probabilities for the time domain patches $\tilde{\mathbf{X}}^{p}$, and $\mathbf{W}_t^p \in \mathbb{R}^{\mathtt{H} \times \mathtt{P}}$ and $\mathbf{W}_{\mathbf{t}} \in \mathbb{R}^{1 \times \mathtt{H}}$ represent the weight matrices of the learnable linear layers.

\textbf{Ensemble strategy:} We adopt a maximum fusion strategy to combine anomaly probabilities from the two branches:
\begin{equation}
  \mathbf{A}^p = \max(\mathbf{A}_f^p, \mathbf{A}_t^p) = [\max(a_{f,1}^p, a_{t,1}^p), \ldots, \max(a_{f,\mathtt{N}}^p, a_{t,\mathtt{N}}^p)],
\end{equation}
where $\mathbf{A}^p\in \mathbb{R}^\mathtt{N}$ represents the ensemble anomaly probabilities for the patches $\mathbf{\tilde{X}}^p$. This strategy ensures that as long as either branch detects an anomaly, it will be retained. It helps reduce false alarms, since both branches must agree on normality. Moreover, it also avoids forcing either branch to fit on detecting anomalies it is less sensitive to, thus improving training stability.
We also explore different ensemble strategies in Section \ref{sec:ablation_study}.

\subsection{Probability-informed Soft Masking MAE}\label{sec:rec}
The core idea of \textsc{CoAD} is to guide MAE reconstruction using prior classification probabilities. We propose a soft masking strategy, where each patch embedding is blended with a learnable mask embedding, weighted by the anomaly probability generated from the classification module.
This enables a more nuanced suppression of potentially anomalous information, especially in borderline cases where a hard mask based on binary classification may be too rigid or error-prone. The resulting soft-masked embeddings are:
\begin{equation}
  \mathbf{E}_{\mathbf{m}}=\mathbf{A}^p \cdot \mathbf{E}_\texttt{mask}+\left(J_{1,\mathtt{N}}-\mathbf{A}^p\right) \cdot \mathbf{W}_{\mathbf{m}}\tilde{\mathbf{X}}^p,
\end{equation}
where $\mathbf{E}_\texttt{mask}\in\mathbb{R}^{\mathtt{H}\times\mathtt{N}}$ represents the learnable mask embedding, $J_{1,\mathtt{N}}$ denotes an all-ones vector of length $\mathtt{N}$, and $\mathbf{W}_{\mathbf{m}} \in \mathbb{R}^{\mathtt{H} \times \mathtt{P}}$ is a learnable projection matrix that maps raw patches into patch embeddings. $\mathbf{E}_{\mathbf{m}}$ is subsequently input into a GRU encoder followed by a linear layer to reconstruct the original time series:
\begin{equation}
  \mathbf{X}_{\mathbf{r}}=\texttt{Flat}(\mathbf{W}_\mathbf{o} \operatorname{GRU}\left(\mathbf{E}_{\mathbf{m}}\right)), \, \text{where}\; \mathbf{W}_\mathbf{o} \in \mathbb{R}^{\mathtt{P} \times \mathtt{H}}.
\end{equation}
The MAE is trained to minimize the Mean Squared Error between the reconstructed sequence $\mathbf{X_r}$ and the original normal time series $\mathbf{X}$, thereby learning to reconstruct normal patterns within masked anomalous regions and to capture only the normal data distribution.

The probability-informed soft masking strategy offers fine-grained control over the masking strength, enabling adaptive suppression of anomalous information based on classification confidence. Unlike rigid hard masking, it handles uncertainty more gracefully, enhancing robustness in ambiguous cases. Moreover, soft masking ensures smoother gradient flow during training and fosters better synergy between the classification and reconstruction branches by aligning their objectives. A comparative analysis of soft- and hard-masking strategies is presented in Section~\ref{sec:ablation_study}.

\subsection{Reconstruction-informed Residual Classification}\label{sec:residual}
Recent advances ~\cite{ResAD2024, adpretrain2025} in image anomaly detection indicate that the discrepancy between the features of anomalous regions and their nearest normal references, referred to as \textbf{residual features}, serves as an intrinsic indicator for anomaly discrimination, exhibiting strong generalization capability in cross-domain settings. The underlying rationale is that, irrespective of the anomaly type, the features of anomalous samples differ substantially from those of the corresponding normal references. Consequently, residual features can be regarded as class-invariant and generalizable representations for anomaly detection.

However, extracting residual features necessitates an extensive reference pool that stores all possible normal features for comparison. Unlike static images, the inherent temporal dynamics of time series data make maintaining a reference pool with sufficient coverage memory-intensive. Furthermore, performing nearest-neighbor retrieval for each input imposes substantial computational overhead in large-scale time series settings. To mitigate these constraints, we propose a simple yet effective approach that directly leverages the MAE reconstruction output as the reference for classification. Since the MAE is optimized to reconstruct normal patterns even within anomalous regions, it provides a ``quasi-normal reference'' for the corresponding input (as visualized in Figure~\ref{fig:insight}). The structure of this reconstruction-informed residual classification module is illustrated in Figure ~\ref{fig:framework}(c), with technical details presented below:

The reconstructed sequence $\mathbf{X_r}$ is first partitioned into patches $\mathbf{X}_\mathbf{r}^{p}$, which are subsequently processed by frequency-branch and time-branch encoders to extract features $\mathbf{H}_f^{p}$ and $\mathbf{H}_t^{p}$, respectively. Residual features $\mathbf{H}_f^{p,r}$ and $\mathbf{H}_t^{p,r}$ are then computed as the differences between the reconstructed patch features and the original input patch features:
\begin{equation}
  \mathbf{H}_f^{p,r} = \mathbf{\tilde{H}}_f^{p} - \mathbf{H}_f^{p}, \, \,
  \mathbf{H}_t^{p,r} = \mathbf{\tilde{H}}_t^{p} - \mathbf{H}_t^{p}.
\end{equation}
The obtained dual-domain residual features are projected through a linear layer followed by a sigmoid activation function to calculate patch-wise anomaly probabilities. These domain-specific probabilities are subsequently aggregated using an element-wise maximum fusion strategy:
\begin{equation}
  \begin{split}
    \mathbf{A}_f^{p,r} & = \sigma\left(\mathbf{W}_{\mathbf{r}}\mathbf{H}_f^{p,r}\right), \, \, \mathbf{A}_t^{p,r} = \sigma\left(\mathbf{W}_{\mathbf{r}}\mathbf{H}_t^{p,r}\right), \\
    \mathbf{A}^{p,r}   & = \max(\mathbf{A}_f^{p,r}, \mathbf{A}_t^{p,r}),
  \end{split}
\end{equation}
where $\mathbf{W}_{\mathbf{r}} \in \mathbb{R}^{1 \times \mathtt{H}}$ denotes a learnable weight vector. The final patch-level anomaly probabilities $\mathbf{A}_c^{p}$ are obtained by averaging the predictions of the time–frequency ensemble classification and residual classification modules:
\begin{equation} \label{equation:average}
  \mathbf{A}_c^{p} = \frac{\mathbf{A}^{p} \oplus \mathbf{A}^{p,r}}{2}, \, \text{where}\;\mathbf{A}_c^p=[a_{c,1}^p, a_{c,2}^p, \ldots, a_{c,\mathtt{N}}^p].
\end{equation}
In Eq.~(\ref{equation:average}), $\mathbf{A}^{p}$ confirms known anomaly patterns, while $\mathbf{A}^{p,r}$ enhances generalization to unseen anomalies. \textbf{The efficacy of the residual classification module in detecting novel anomalies is experimentally validated in Appendix \ref{sec:generalization}.}

\subsection{Cooperative Training and Joint Anomaly Inference}\label{sec:train}

\subsubsection{Cooperative Training} The classification and reconstruction modules are jointly trained in an end-to-end manner.
The overall training objective integrates the binary cross-entropy (BCE) loss for classification and the mean squared error (MSE) loss for reconstruction:
{
\small
\begin{equation}
  \begin{split}
    \mathcal{L}_{BCE} & = \frac{1}{N}\sum_{i=1}^{N}\left(y_i \cdot \log(a_{c,i}^{p}) + (1-y_i) \cdot \log(1-a_{c,i}^{p})\right), \\
    \mathcal{L}_{MSE} & = \frac{1}{N}\sum_{i=1}^{N}\left\| x_{r,i}^{p} - x_{i}^{p} \right\|_{2}^{2}, \, \,
    \mathcal{L}_{Final} = \mathcal{L}_{BCE} + \lambda \cdot \mathcal{L}_{MSE},
  \end{split}
\end{equation}
}
where $a_{c,i}^{p}$ denotes the anomaly probability of the patch $\tilde{x}^p_{i}$, and $y_{i}$ is its corresponding label, where $y_{i} =1$ if the patch contains anomalies and $y_{i}=0$ otherwise. $x_{r,i}^p$ is the reconstructed patch and $x_{i}^p$ is the initial normal time series patch. $\lambda$ is the weight to balance the two losses.

\subsubsection{Joint Anomaly Inference} During inference, the classification and reconstruction modules operate collaboratively to enable comprehensive anomaly detection. The final anomaly score for each patch is computed as the sum of the anomaly probability and the reconstruction error:
\begin{equation}
  a(x_{i}^p) = a_{c,i}^{p} \cdot J_{1,\mathtt{P}} + \left| \tilde{x}^p_{i} - x_{r,i}^p \right|,
\end{equation}
where $J_{1,\mathtt{P}}$ denotes an all-ones vector of length $\mathtt{P}$, ensuring proper dimensional alignment for element-wise addition. Since the input time series is normalized using the training set, both the classification probabilities and reconstruction errors are on comparable scales, allowing them to be directly added into a unified anomaly score. Following prior studies \cite{tang2025MMMA, qinMemoryAugmentedUTransformerMultivariate2023, sunUnravelingAnomalyTime2024}, a moving average is applied to smooth the unified anomaly scores.


\section{Experiments}\label{sec:experiment}

\subsection{Datasets and Evaluation Metrics}
\subsubsection{Current Issues} Unreliable datasets and biased evaluation metrics have long plagued the field of time series anomaly detection \cite{liu2024the, wuCurrentTimeSeries2021}.
Many studies reported impressive performance based on flawed datasets and metrics, thereby contributing to the problem of \emph{``Creating the Illusion of Progress''} \cite{wuCurrentTimeSeries2021,eamonnKeoghIrrational21}.
Commonly used datasets such as SMD \cite{suRobustAnomalyDetection2019}, PSM \cite{abdulaalPracticalApproachAsynchronous2021}, SWaT \cite{gohDatasetSupportResearch2017}, SMAP \cite{hundmanDetectingSpacecraftAnomalies2018}, MSL \cite{hundmanDetectingSpacecraftAnomalies2018}, and NAB \cite{AHMAD2017134}, suffer from several critical issues including mislabeled ground truth, trivial anomalies, unrealistic anomaly densities, and the run-to-failure bias \cite{liu2024the,wuCurrentTimeSeries2021, eamonnKeoghIrrational21, 2024Fundamental}.
Moreover, some evaluation metrics, such as point-adjusted F1 (F1-PA)~\cite{suRobustAnomalyDetection2019} and F1-Affiliation \cite{huetLocalEvaluationTime2022}, tend to overestimate model performance, often awarding the highest scores to random predictions~\cite{tang2025MMMA, liu2024the, kimRigorousEvaluationTimeSeries2022}.

\subsubsection{Our Settings}
To ensure the reliability of our evaluation, we adopt recently proposed, largest-scale and highest-quality datasets \cite{liu2024the, wuCurrentTimeSeries2021}, along with the most rigorous evaluation metrics \cite{paparrizosVolumeSurfaceNew2022}.

\underline{\textit{For datasets}}, we use the KDD21 \cite{wuCurrentTimeSeries2021} and TSB-AD \cite{liu2024the} benchmarks. \textit{i)} KDD21 is widely acknowledged for its high data quality and domain diversity, comprising 250 datasets across various fields, including healthcare, sports, industry, and robotics. \textit{ii)} TSB-AD also covers various domains and partially overlaps with KDD21.
However, certain datasets within TSB-AD, such as NAB, WSD, YAHOO, and Stock, still suffer from the aforementioned data quality issues \cite{2024Fundamental}. Therefore, following established dataset quality criteria \cite{schmidlAnomalyDetectionTime2022, wuCurrentTimeSeries2021, luDAMPAccurateTime2023}, we select only the high-quality, non-overlapping datasets from TSB-AD for our experiments. In total, our evaluation encompasses \textbf{314 datasets} from diverse domains drawn from these two benchmarks, ensuring a comprehensive assessment.  \textbf{Detailed dataset descriptions are provided in Appendix \ref{sec:datasets}.}

\underline{\textit{For metrics}}, we adopt the evaluation metrics recommended by recent benchmarking studies \cite{liu2024the, paparrizosVolumeSurfaceNew2022, vus2025Boniol}, including Standard-F1, AUC-PR, Range-AUC-PR, and VUS-PR, which are recognized as the most reliable and precise measures in current research \cite{liu2024the}. \textbf{Details of these metrics are available in Appendix \ref{sec:metrics}.}

\subsection{Baselines and Implementation Details} \label{sec:baselines}
\subsubsection{Baselines}
To demonstrate the superiority of our method, we compared it against \textbf{24 SOTA} baseline methods, including 17 deep learning-based and 7 data mining-based methods. As shown in Table \ref{tab:main_result}, the deep learning-based methods can be categorized into four groups:\ul{\itshape i) Pure MAE-based methods} and \underline{\itshape ii) Pure OE-based methods} are standalone MAE and OE approaches; \ul{\itshape iii) Time-Frequency Reconstruction methods} perform reconstruction in both time and frequency domains; and \ul{\itshape iv) Other Deep Learning-based methods} rely on non-masking reconstruction (TranAD, MAUT and FITS), forecasting (M2N2), or feature discrepancy (DCdetector and AnomalyTransformer). Notably, DADA and MOMENT are built on large foundation models.

\subsubsection{Implementation Details}
Both our model and all baselines follow identical data preprocessing procedures within an integrated, unified pipeline. \textbf{More implementation details of \textsc{CoAD} and baselines are available in Appendix \ref{sec:implementation}.}

\subsection{Comparison results}
\begin{table*}[!t]
  \caption{Average results (\%) on KDD21 and TSB-AD. The best results are in \textbf{bold}, and the second-best results are with \underline{underline}.}
  \label{tab:main_result}
  \renewcommand{\arraystretch}{0.85}
  \resizebox{0.9\textwidth}{!}{%
    \begin{tabular}{@{}cccccccccc@{}}
      \toprule
      \multicolumn{2}{c|}{\textbf{Model / Dataset}}                                                                  &
      \multicolumn{4}{c|}{\textbf{KDD21}}                                                                            &
      \multicolumn{4}{c}{\textbf{TSB-AD}}                                                                              \\ \midrule
      \multicolumn{1}{c|}{Model Class}                                                                               &
      \multicolumn{1}{c|}{Model (Venue)}                                                                             &
      F1                                                                                                             &
      AUC-PR                                                                                                         &
      R-AUC-PR                                                                                                       &
      \multicolumn{1}{c|}{VUS-PR}                                                                                    &
      F1                                                                                                             &
      AUC-PR                                                                                                         &
      R-AUC-PR                                                                                                       &
      VUS-PR                                                                                                           \\ \midrule
      \multicolumn{1}{c|}{}                                                                                          &
      \multicolumn{1}{c|}{DADA (ICLR-25 \cite{shentu2025towards})}                                                   &
      3.49                                                                                                           &
      1.39                                                                                                           &
      2.04                                                                                                           &
      \multicolumn{1}{c|}{2.05}                                                                                      &
      17.36                                                                                                          &
      12.30                                                                                                          &
      14.92                                                                                                          &
      14.53                                                                                                            \\
      \multicolumn{1}{c|}{}                                                                                          &
      \multicolumn{1}{c|}{MOMENT (ICML-24 \cite{goswami2024moment})}                                                 &
      11.06                                                                                                          &
      7.71                                                                                                           &
      9.20                                                                                                           &
      \multicolumn{1}{c|}{9.13}                                                                                      &
      25.29                                                                                                          &
      18.80                                                                                                          &
      25.90                                                                                                          &
      24.67                                                                                                            \\
      \multicolumn{1}{c|}{}                                                                                          &
      \multicolumn{1}{c|}{MMA (VLDB-25 \cite{tang2025MMMA})}                                                         &
      \ul{44.47}                                                                                                     &
      39.24                                                                                                          &
      37.97                                                                                                          &
      \multicolumn{1}{c|}{37.38}                                                                                     &
      \ul{43.20}                                                                                                     &
      \ul{38.65}                                                                                                     &
      37.33                                                                                                          &
      \ul{36.94}                                                                                                       \\
      \multicolumn{1}{c|}{\multirow{-4}{*}{Pure MAE}}                                                                &
      \multicolumn{1}{c|}{TFMAE (ICDE-24 \cite{fangTemporalFrequencyMaskedAutoencoders2024})}                        &
      2.53                                                                                                           &
      0.99                                                                                                           &
      1.75                                                                                                           &
      \multicolumn{1}{c|}{1.75}                                                                                      &
      7.29                                                                                                           &
      3.45                                                                                                           &
      5.83                                                                                                           &
      5.77                                                                                                             \\ \midrule
      \multicolumn{1}{c|}{}                                                                                          &
      \multicolumn{1}{c|}{AnomalyBERT (ICLR-23 \cite{jeongAnomalyBERTSelfSupervisedTransformer2023})}                &
      23.42                                                                                                          &
      15.92                                                                                                          &
      14.16                                                                                                          &
      \multicolumn{1}{c|}{13.79}                                                                                     &
      19.14                                                                                                          &
      9.85                                                                                                           &
      13.21                                                                                                          &
      13.85                                                                                                            \\
      \multicolumn{1}{c|}{}                                                                                          &
      \multicolumn{1}{c|}{CutAddPaste (KDD-24 \cite{wangCutAddPasteTimeSeries2024})}                                 &
      21.75                                                                                                          &
      15.51                                                                                                          &
      19.00                                                                                                          &
      \multicolumn{1}{c|}{18.20}                                                                                     &
      26.22                                                                                                          &
      21.34                                                                                                          &
      25.45                                                                                                          &
      25.08                                                                                                            \\
      \multicolumn{1}{c|}{}                                                                                          &
      \multicolumn{1}{c|}{TriAD (ICDE-24 \cite{sunUnravelingAnomalyTime2024})}                                       &
      16.78                                                                                                          &
      22.37                                                                                                          &
      28.09                                                                                                          &
      \multicolumn{1}{c|}{27.56}                                                                                     &
      N/A                                                                                                            &
      N/A                                                                                                            &
      N/A                                                                                                            &
      N/A                                                                                                              \\
      \multicolumn{1}{c|}{\multirow{-4}{*}{Pure OE}}                                                                 &
      \multicolumn{1}{c|}{COUTA (TKDE-24 \cite{xuCalibratedOneclassClassification2024})}                             &
      6.58                                                                                                           &
      3.65                                                                                                           &
      3.79                                                                                                           &
      \multicolumn{1}{c|}{3.82}                                                                                      &
      18.78                                                                                                          &
      13.12                                                                                                          &
      11.07                                                                                                          &
      11.01                                                                                                            \\ \midrule
      \multicolumn{1}{c|}{}                                                                                          &
      \multicolumn{1}{c|}{FCVAE (WWW-24 \cite{wangRevisitingVAEUnsupervised2024a})}                                  &
      11.23                                                                                                          &
      7.18                                                                                                           &
      6.25                                                                                                           &
      \multicolumn{1}{c|}{6.38}                                                                                      &
      28.47                                                                                                          &
      22.12                                                                                                          &
      20.61                                                                                                          &
      20.58                                                                                                            \\
      \multicolumn{1}{c|}{}                                                                                          &
      \multicolumn{1}{c|}{TFAD (CIKM-22 \cite{zhangTFADDecompositionTime2022})}                                      &
      1.85                                                                                                           &
      0.86                                                                                                           &
      1.59                                                                                                           &
      \multicolumn{1}{c|}{1.58}                                                                                      &
      6.17                                                                                                           &
      3.34                                                                                                           &
      6.21                                                                                                           &
      5.99                                                                                                             \\
      \multicolumn{1}{c|}{\multirow{-3}{*}{\begin{tabular}[c]{@{}c@{}}Time-Frequency\\ Reconstruction\end{tabular}}} &
      \multicolumn{1}{c|}{CATCH (ICLR-25 \cite{wu2024catch})}                                                        &
      13.29                                                                                                          &
      9.08                                                                                                           &
      9.26                                                                                                           &
      \multicolumn{1}{c|}{9.19}                                                                                      &
      24.46                                                                                                          &
      20.02                                                                                                          &
      22.32                                                                                                          &
      21.67                                                                                                            \\ \midrule
      \multicolumn{1}{c|}{}                                                                                          &
      \multicolumn{1}{c|}{TranAD (VLDB-22 \cite{tuliTranADDeepTransformer2022})}                                     &
      11.23                                                                                                          &
      7.78                                                                                                           &
      7.94                                                                                                           &
      \multicolumn{1}{c|}{7.89}                                                                                      &
      18.07                                                                                                          &
      13.06                                                                                                          &
      12.23                                                                                                          &
      12.05                                                                                                            \\
      \multicolumn{1}{c|}{}                                                                                          &
      \multicolumn{1}{c|}{MAUT (ICASSP-23 \cite{qinMemoryAugmentedUTransformerMultivariate2023})}                    &
      30.20                                                                                                          &
      23.84                                                                                                          &
      23.94                                                                                                          &
      \multicolumn{1}{c|}{23.65}                                                                                     &
      19.82                                                                                                          &
      14.67                                                                                                          &
      14.91                                                                                                          &
      14.75                                                                                                            \\
      \multicolumn{1}{c|}{}                                                                                          &
      \multicolumn{1}{c|}{M2N2 (AAAI-24 \cite{kimWhenModelMeets2024})}                                               &
      5.57                                                                                                           &
      2.70                                                                                                           &
      3.15                                                                                                           &
      \multicolumn{1}{c|}{3.18}                                                                                      &
      16.98                                                                                                          &
      10.38                                                                                                          &
      9.41                                                                                                           &
      9.25                                                                                                             \\
      \multicolumn{1}{c|}{}                                                                                          &
      \multicolumn{1}{c|}{FITS (ICLR-24 \cite{xu2024fits})}                                                          &
      5.34                                                                                                           &
      2.63                                                                                                           &
      3.46                                                                                                           &
      \multicolumn{1}{c|}{3.45}                                                                                      &
      13.23                                                                                                          &
      7.14                                                                                                           &
      10.63                                                                                                          &
      10.37                                                                                                            \\
      \multicolumn{1}{c|}{}                                                                                          &
      \multicolumn{1}{c|}{DCdetector (KDD-23 \cite{yang2023DCdetector})}                                             &
      2.66                                                                                                           &
      1.09                                                                                                           &
      1.88                                                                                                           &
      \multicolumn{1}{c|}{1.85}                                                                                      &
      6.47                                                                                                           &
      3.12                                                                                                           &
      5.79                                                                                                           &
      5.69                                                                                                             \\
      \multicolumn{1}{c|}{\multirow{-6}{*}{Other Deep Learning}}                                                     &
      \multicolumn{1}{c|}{AnomalyTrans (ICLR-22 \cite{xu2022anomaly})}                                               &
      2.37                                                                                                           &
      0.97                                                                                                           &
      1.87                                                                                                           &
      \multicolumn{1}{c|}{1.80}                                                                                      &
      7.01                                                                                                           &
      3.36                                                                                                           &
      6.27                                                                                                           &
      6.13                                                                                                             \\ \midrule
      \multicolumn{1}{c|}{}                                                                                          &
      \multicolumn{1}{c|}{KShapeAD (NeurIPS-24 \cite{liu2024the})}                                                   &
      43.27                                                                                                          &
      \ul{39.37}                                                                                                     &
      \ul{39.48}                                                                                                     &
      \multicolumn{1}{c|}{\ul{39.02}}                                                                                &
      35.51                                                                                                          &
      32.67                                                                                                          &
      31.87                                                                                                          &
      31.27                                                                                                            \\
      \multicolumn{1}{c|}{}                                                                                          &
      \multicolumn{1}{c|}{SAND (VLDB-21 \cite{boniolSANDStreamingSubsequence2021})}                                  &
      39.43                                                                                                          &
      34.72                                                                                                          &
      34.23                                                                                                          &
      \multicolumn{1}{c|}{33.78}                                                                                     &
      \multicolumn{1}{l}{34.74}                                                                                      &
      31.52                                                                                                          &
      31.85                                                                                                          &
      31.05                                                                                                            \\
      \multicolumn{1}{c|}{}                                                                                          &
      \multicolumn{1}{c|}{Sub-PCA (NeurIPS-24 \cite{liu2024the})}                                                    &
      15.45                                                                                                          &
      11.32                                                                                                          &
      14.06                                                                                                          &
      \multicolumn{1}{c|}{13.17}                                                                                     &
      32.49                                                                                                          &
      27.59                                                                                                          &
      23.88                                                                                                          &
      23.85                                                                                                            \\
      \multicolumn{1}{c|}{}                                                                                          &
      \multicolumn{1}{c|}{Series2Graph (VLDB-20 \cite{boniolSeries2GraphGraphbasedSubsequence2020})}                 &
      28.11                                                                                                          &
      22.61                                                                                                          &
      25.63                                                                                                          &
      \multicolumn{1}{c|}{24.81}                                                                                     &
      33.48                                                                                                          &
      30.11                                                                                                          &
      30.13                                                                                                          &
      29.57                                                                                                            \\
      \multicolumn{1}{c|}{}                                                                                          &
      \multicolumn{1}{c|}{KMeansAD (VLDB-22 \cite{yairi2001fault})}                                                  &
      37.97                                                                                                          &
      34.33                                                                                                          &
      33.49                                                                                                          &
      \multicolumn{1}{c|}{33.20}                                                                                     &
      41.42                                                                                                          &
      37.26                                                                                                          &
      \ul{37.64}                                                                                                     &
      36.81                                                                                                            \\
      \multicolumn{1}{c|}{}                                                                                          &
      \multicolumn{1}{c|}{Matrix Profile (CIKM-16 \cite{nakamuraMERLINParameterFreeDiscovery2020})}                  &
      28.00                                                                                                          &
      18.50                                                                                                          &
      25.37                                                                                                          &
      \multicolumn{1}{c|}{24.13}                                                                                     &
      35.10                                                                                                          &
      27.93                                                                                                          &
      29.88                                                                                                          &
      28.94                                                                                                            \\
      \multicolumn{1}{c|}{\multirow{-7}{*}{Data Mining}}                                                             &
      \multicolumn{1}{c|}{DAMP (KDD-22 \cite{luDAMPAccurateTime2023} )}                                              &
      29.31                                                                                                          &
      18.99                                                                                                          &
      25.09                                                                                                          &
      \multicolumn{1}{c|}{24.18}                                                                                     &
      17.44                                                                                                          &
      11.59                                                                                                          &
      12.98                                                                                                          &
      12.60                                                                                                            \\ \midrule
      \rowcolor[HTML]{EFEFEF}
      \multicolumn{1}{c|}{\cellcolor[HTML]{EFEFEF}\textbf{Cooperative}}                                              &
      \multicolumn{1}{c|}{\cellcolor[HTML]{EFEFEF}\textbf{\textsc{CoAD} (ours)}}                                     &
      \textbf{52.82}                                                                                                 &
      \textbf{48.10}                                                                                                 &
      \textbf{46.35}                                                                                                 &
      \multicolumn{1}{c|}{\cellcolor[HTML]{EFEFEF}\textbf{45.67}}                                                    &
      \textbf{49.13}                                                                                                 &
      \textbf{43.66}                                                                                                 &
      \textbf{40.50}                                                                                                 &
      \textbf{39.83}                                                                                                   \\ \bottomrule
    \end{tabular}%
  }
  \vspace{-0.3cm}
\end{table*}

\subsubsection{Effectiveness} \label{sec:effectiveness}
The comparison results are summarized in Table~\ref{tab:main_result} (\textbf{results with KDD21 cup metrics are in Appendex \ref{sec:KDD21_score}}). The following key observations can be made:
1) \textsc{CoAD} consistently surpasses all baseline models on both the KDD21 and TSB-AD benchmarks across all evaluation metrics.
2) Compared with methods that rely exclusively on either MAE or OE, \textsc{CoAD} achieves substantial performance improvements, demonstrating the effectiveness of the proposed cooperative framework in leveraging the complementary strengths of classification and reconstruction.
3) \textsc{CoAD} also outperforms other OE-based methods that incorporate a broader range of simulated anomaly types during training. This highlights the superior generalization capabilities of \textsc{CoAD}, delivering stronger detection performance while requiring far less prior knowledge.
4) Notably, data mining-based methods outperform many deep learning counterparts, consistent with prior findings \cite{schmidlAnomalyDetectionTime2022,laiRevisitingTimeSeries2021,rewickiItWorthIt2023,liu2024the, Position24, mejriUnsupervisedAnomalyDetection2024,gargEvaluationAnomalyDetection2022}, highlighting the ongoing challenges of deep learning in time series anomaly detection. Nevertheless, \textsc{CoAD} surpasses even the strongest data mining baselines, illustrating that with principled architectural design and rigorous evaluation, deep learning can achieve SOTA performance in TSAD. \textbf{Appendix \ref{sec:parameter} further investigates the influence of various hyperparameters, including the window size $\mathtt{T}$, patch size $\mathtt{P}$, loss weight $\lambda$, and different encoder backbone choices (GRU or Transformer) on model effectiveness.}

\begin{figure}[t]
  \centering
  \includegraphics[width=\columnwidth]{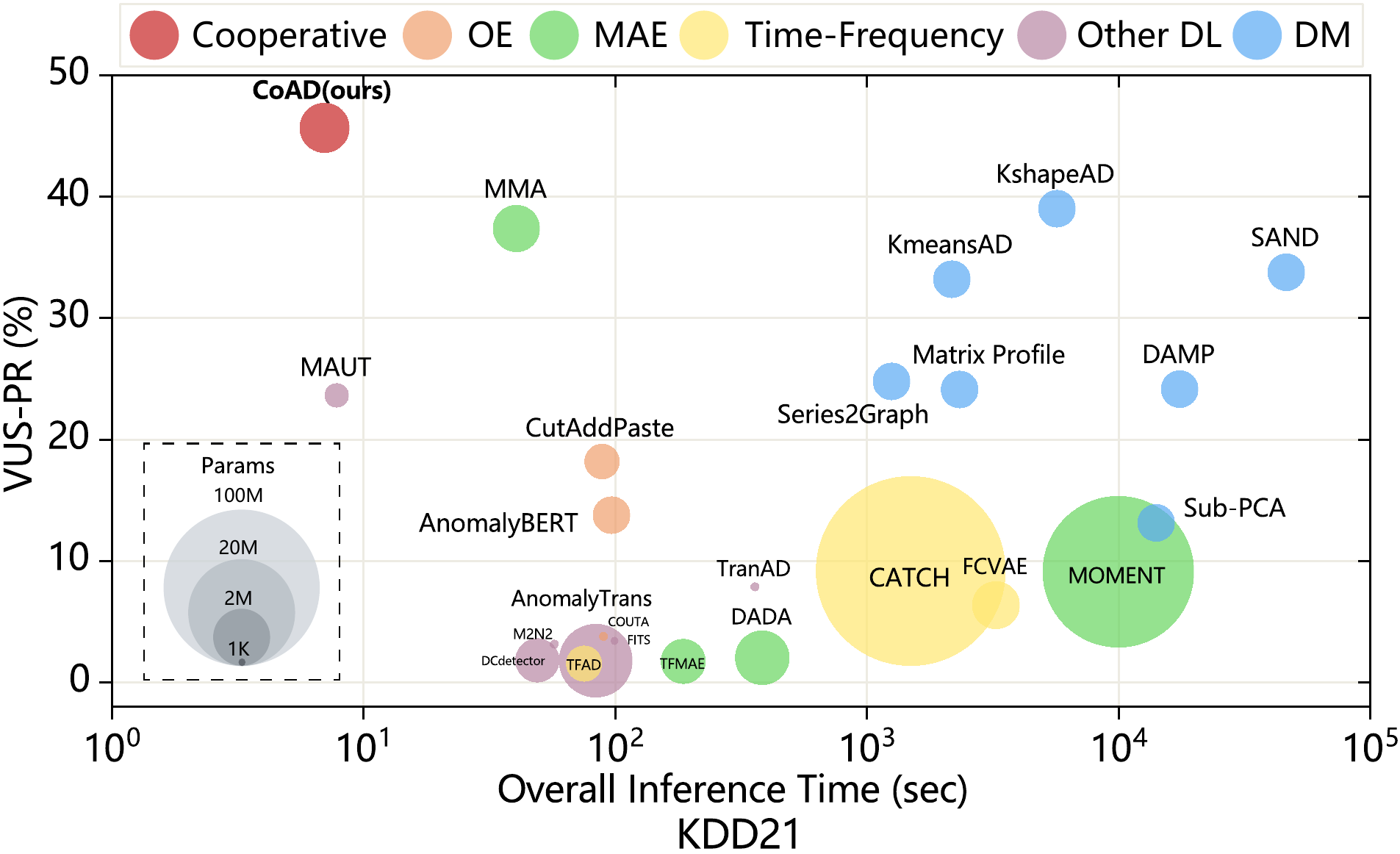}
  \caption{Model efficiency comparison.}
  \label{fig:KDD21_efficiency}
  \vspace{-0.5cm}
\end{figure}

\subsubsection{Efficiency} \label{sec:efficiency}
Figure~\ref{fig:KDD21_efficiency} presents a comparison of model efficiency in terms of overall inference time and parameter count on the KDD21 benchmark (\textbf{the settings and results on the TSB-AD benchmark are available in Appendix \ref{sec:tsb_ad_efficiency}}). The results highlight that \textsc{CoAD} not only achieves superior detection performance but also delivers remarkable computational efficiency.
Specifically, \textsc{CoAD} completes inference over the entire KDD21 benchmark (with dataset lengths ranging from \textbf{6K} to \textbf{650K} data points), comprising more than \textbf{6 million} data points in total, in only \textbf{6.89 seconds}, outperforming most baselines by orders of magnitude in speed. This underscores the strong potential of \textsc{CoAD} for real-time anomaly detection in high-throughput data streams.
We also evaluate training efficiency on entity 241, the largest dataset in the KDD21 benchmark, which contains \textbf{250K} training samples. \textsc{CoAD} achieves a training speed of \textbf{0.4066 seconds per epoch}, confirming its high training efficiency.
Furthermore, the model maintains a compact size of only \textbf{2.04M} parameters, highlighting its practicality for deployment in resource-constrained environments. \textbf{The theoretical time complexity analysis is provided in Appendix \ref{sec:time_complex}.}

\subsection{Ablation and Design Choice Study}
\label{sec:ablation_study}

\begin{table*}[t]
  \caption{Quantitative ablation results. Best results are in \textbf{bold}.}
  \label{tab:ablation_study}
  \resizebox{0.9\textwidth}{!}{%
    \begin{tabular}{@{}ccccccccccccccc@{}}
      \toprule
      \multicolumn{1}{c|}{}                                                             &
      \multicolumn{2}{c|}{\textbf{OE}}                                                  &
      \multicolumn{4}{c|}{\textbf{MAE}}                                                 &
      \multicolumn{4}{c|}{\textbf{KDD21}}                                               &
      \multicolumn{4}{c}{\textbf{TSB-AD}}                                                 \\ \cmidrule(l){2-15}
      \multicolumn{1}{c|}{\multirow{-2}{*}{\textbf{Variants}}}                          &
      Time                                                                              &
      \multicolumn{1}{c|}{Frequency}                                                    &
      Random                                                                            &
      Grating                                                                           &
      \begin{tabular}[c]{@{}c@{}}Guide \\ w/ Hard Mask\end{tabular}                     &
      \multicolumn{1}{c|}{\begin{tabular}[c]{@{}c@{}}Guide\\ w/ Soft Mask\end{tabular}} &
      F1                                                                                &
      AUC-PR                                                                            &
      R-AUC-PR                                                                          &
      \multicolumn{1}{c|}{VUS-PR}                                                       &
      F1                                                                                &
      AUC-PR                                                                            &
      R-AUC-PR                                                                          &
      VUS-PR                                                                              \\ \midrule
      \multicolumn{1}{c|}{}                                                             &
      \Checkmark                                                                        &
      \multicolumn{1}{c|}{\textendash}                                                  &
      \textendash                                                                       &
      \textendash                                                                       &
      \textendash                                                                       &
      \multicolumn{1}{c|}{\textendash}                                                  &
      27.84                                                                             &
      24.12                                                                             &
      24.81                                                                             &
      \multicolumn{1}{c|}{24.38}                                                        &
      29.79                                                                             &
      22.36                                                                             &
      27.14                                                                             &
      26.28                                                                               \\
      \multicolumn{1}{c|}{\multirow{-2}{*}{\textbf{OE alone}}}                          &
      \Checkmark                                                                        &
      \multicolumn{1}{c|}{\Checkmark}                                                   &
      \textendash                                                                       &
      \textendash                                                                       &
      \textendash                                                                       &
      \multicolumn{1}{c|}{\textendash}                                                  &
      47.95                                                                             &
      42.75                                                                             &
      41.06                                                                             &
      \multicolumn{1}{c|}{40.41}                                                        &
      37.68                                                                             &
      31.21                                                                             &
      30.48                                                                             &
      29.90                                                                               \\ \midrule
      \multicolumn{1}{c|}{}                                                             &
      \textendash                                                                       &
      \multicolumn{1}{c|}{\textendash}                                                  &
      \Checkmark                                                                        &
      \textendash                                                                       &
      \textendash                                                                       &
      \multicolumn{1}{c|}{\textendash}                                                  &
      35.65                                                                             &
      30.23                                                                             &
      30.96                                                                             &
      \multicolumn{1}{c|}{30.24}                                                        &
      32.30                                                                             &
      25.74                                                                             &
      23.55                                                                             &
      23.35                                                                               \\
      \multicolumn{1}{c|}{\multirow{-2}{*}{\textbf{MAE alone}}}                         &
      \textendash                                                                       &
      \multicolumn{1}{c|}{\textendash}                                                  &
      \textendash                                                                       &
      \Checkmark                                                                        &
      \textendash                                                                       &
      \multicolumn{1}{c|}{\textendash}                                                  &
      38.99                                                                             &
      32.31                                                                             &
      32.76                                                                             &
      \multicolumn{1}{c|}{31.99}                                                        &
      35.01                                                                             &
      29.08                                                                             &
      28.51                                                                             &
      28.03                                                                               \\ \midrule
      \multicolumn{1}{c|}{}                                                             &
      \Checkmark                                                                        &
      \multicolumn{1}{c|}{\Checkmark}                                                   &
      \Checkmark                                                                        &
      \textendash                                                                       &
      \textendash                                                                       &
      \multicolumn{1}{c|}{\textendash}                                                  &
      44.88                                                                             &
      41.26                                                                             &
      39.55                                                                             &
      \multicolumn{1}{c|}{39.05}                                                        &
      45.01                                                                             &
      39.31                                                                             &
      35.79                                                                             &
      35.20                                                                               \\
      \multicolumn{1}{c|}{}                                                             &
      \Checkmark                                                                        &
      \multicolumn{1}{c|}{\Checkmark}                                                   &
      \textendash                                                                       &
      \Checkmark                                                                        &
      \textendash                                                                       &
      \multicolumn{1}{c|}{\textendash}                                                  &
      46.16                                                                             &
      41.95                                                                             &
      40.86                                                                             &
      \multicolumn{1}{c|}{40.19}                                                        &
      47.07                                                                             &
      41.48                                                                             &
      37.39                                                                             &
      37.08                                                                               \\
      \multicolumn{1}{c|}{}                                                             &
      \Checkmark                                                                        &
      \multicolumn{1}{c|}{\Checkmark}                                                   &
      \textendash                                                                       &
      \textendash                                                                       &
      \Checkmark                                                                        &
      \multicolumn{1}{c|}{\textendash}                                                  &
      49.03                                                                             &
      45.27                                                                             &
      43.87                                                                             &
      \multicolumn{1}{c|}{43.49}                                                        &
      44.06                                                                             &
      38.57                                                                             &
      35.31                                                                             &
      34.76                                                                               \\
      \multicolumn{1}{c|}{\multirow{-4}{*}{\textbf{Cooperative}}}                       &
      \Checkmark                                                                        &
      \multicolumn{1}{c|}{\textendash}                                                  &
      \textendash                                                                       &
      \textendash                                                                       &
      \textendash                                                                       &
      \multicolumn{1}{c|}{\Checkmark}                                                   &
      48.45                                                                             &
      43.21                                                                             &
      42.87                                                                             &
      \multicolumn{1}{c|}{42.35}                                                        &
      43.10                                                                             &
      36.87                                                                             &
      37.91                                                                             &
      36.92                                                                               \\ \midrule
      \rowcolor[HTML]{EFEFEF}
      \multicolumn{1}{c|}{\cellcolor[HTML]{EFEFEF}\textbf{CoAD}}                        &
      \Checkmark                                                                        &
      \multicolumn{1}{c|}{\cellcolor[HTML]{EFEFEF}\Checkmark}                           &
      \textendash                                                                       &
      \textendash                                                                       &
      \textendash                                                                       &
      \multicolumn{1}{c|}{\cellcolor[HTML]{EFEFEF}\Checkmark}                           &
      \textbf{52.82}                                                                    &
      \textbf{48.10}                                                                    &
      \textbf{46.35}                                                                    &
      \multicolumn{1}{c|}{\cellcolor[HTML]{EFEFEF}\textbf{45.67}}                       &
      \textbf{49.13}                                                                    &
      \textbf{43.66}                                                                    &
      \textbf{40.50}                                                                    &
      \textbf{39.83}                                                                      \\ \midrule
      \multicolumn{15}{c}{}                                                               \\
      \multicolumn{15}{c}{\multirow{-2}{*}{\textbf{Design Choice}}}                       \\ \midrule
      \multicolumn{1}{c|}{\textbf{Variants}}                                            &
      \multicolumn{2}{c|}{\textbf{Classification Granularity}}                          &
      \multicolumn{3}{c|}{\textbf{Fusion Strategy}}                                     &
      \multicolumn{1}{c|}{\textbf{Masking Method}}                                      &
      \multicolumn{4}{c|}{\textbf{KDD21}}                                               &
      \multicolumn{4}{c}{\textbf{TSB-AD}}                                                 \\ \midrule
      \multicolumn{1}{c|}{}                                                             &
      Step Level                                                                        &
      \multicolumn{1}{c|}{Window Level}                                                 &
      Feature\_Add                                                                      &
      Feature\_Gate                                                                     &
      \multicolumn{1}{c|}{Decision\_Mean}                                               &
      \multicolumn{1}{c|}{\begin{tabular}[c]{@{}c@{}}Guide \\ w/o Score\end{tabular}}   &
      F1                                                                                &
      AUC-PR                                                                            &
      R-AUC-PR                                                                          &
      \multicolumn{1}{c|}{VUS-PR}                                                       &
      F1                                                                                &
      AUC-PR                                                                            &
      R-AUC-PR                                                                          &
      VUS-PR                                                                              \\ \cmidrule(l){2-15}
      \multicolumn{1}{c|}{}                                                             &
      \Checkmark                                                                        &
      \multicolumn{1}{c|}{\textendash}                                                  &
      \textendash                                                                       &
      \textendash                                                                       &
      \multicolumn{1}{c|}{\textendash}                                                  &
      \multicolumn{1}{c|}{\textendash}                                                  &
      43.09                                                                             &
      36.62                                                                             &
      35.43                                                                             &
      \multicolumn{1}{c|}{35.00}                                                        &
      42.52                                                                             &
      36.06                                                                             &
      34.02                                                                             &
      33.37                                                                               \\
      \multicolumn{1}{c|}{}                                                             &
      \textendash                                                                       &
      \multicolumn{1}{c|}{\Checkmark}                                                   &
      \textendash                                                                       &
      \textendash                                                                       &
      \multicolumn{1}{c|}{\textendash}                                                  &
      \multicolumn{1}{c|}{\textendash}                                                  &
      15.26                                                                             &
      10.21                                                                             &
      11.29                                                                             &
      \multicolumn{1}{c|}{11.02}                                                        &
      22.43                                                                             &
      17.08                                                                             &
      15.71                                                                             &
      15.53                                                                               \\
      \multicolumn{1}{c|}{}                                                             &
      \textendash                                                                       &
      \multicolumn{1}{c|}{\textendash}                                                  &
      \Checkmark                                                                        &
      \textendash                                                                       &
      \multicolumn{1}{c|}{\textendash}                                                  &
      \multicolumn{1}{c|}{\textendash}                                                  &
      50.76                                                                             &
      45.93                                                                             &
      44.76                                                                             &
      \multicolumn{1}{c|}{44.25}                                                        &
      40.11                                                                             &
      35.15                                                                             &
      33.40                                                                             &
      32.94                                                                               \\
      \multicolumn{1}{c|}{}                                                             &
      \textendash                                                                       &
      \multicolumn{1}{c|}{\textendash}                                                  &
      \textendash                                                                       &
      \Checkmark                                                                        &
      \multicolumn{1}{c|}{\textendash}                                                  &
      \multicolumn{1}{c|}{\textendash}                                                  &
      50.52                                                                             &
      45.58                                                                             &
      45.53                                                                             &
      \multicolumn{1}{c|}{44.71}                                                        &
      42.23                                                                             &
      37.20                                                                             &
      34.87                                                                             &
      34.40                                                                               \\
      \multicolumn{1}{c|}{}                                                             &
      \textendash                                                                       &
      \multicolumn{1}{c|}{\textendash}                                                  &
      \textendash                                                                       &
      \textendash                                                                       &
      \multicolumn{1}{c|}{\Checkmark}                                                   &
      \multicolumn{1}{c|}{\textendash}                                                  &
      51.50                                                                             &
      45.98                                                                             &
      45.53                                                                             &
      \multicolumn{1}{c|}{44.81}                                                        &
      42.24                                                                             &
      36.23                                                                             &
      33.71                                                                             &
      33.18                                                                               \\
      \multicolumn{1}{c|}{\multirow{-7}{*}{\textbf{Cooperative}}}                       &
      \textendash                                                                       &
      \multicolumn{1}{c|}{\textendash}                                                  &
      \textendash                                                                       &
      \textendash                                                                       &
      \multicolumn{1}{c|}{\textendash}                                                  &
      \multicolumn{1}{c|}{\Checkmark}                                                   &
      48.13                                                                             &
      42.92                                                                             &
      42.93                                                                             &
      \multicolumn{1}{c|}{42.09}                                                        &
      37.07                                                                             &
      30.54                                                                             &
      28.04                                                                             &
      27.73                                                                               \\ \midrule
      \rowcolor[HTML]{EFEFEF}
      \multicolumn{1}{c|}{\cellcolor[HTML]{EFEFEF}\textbf{CoAD}}                        &
      \multicolumn{2}{c|}{\cellcolor[HTML]{EFEFEF}\textbf{Patch Level}}                 &
      \multicolumn{3}{c|}{\cellcolor[HTML]{EFEFEF}\textbf{Decision\_Max}}               &
      \multicolumn{1}{c|}{\cellcolor[HTML]{EFEFEF}\textbf{Guide w/ Score}}              &
      \textbf{52.82}                                                                    &
      \textbf{48.10}                                                                    &
      \textbf{46.35}                                                                    &
      \multicolumn{1}{c|}{\cellcolor[HTML]{EFEFEF}\textbf{45.67}}                       &
      \cellcolor[HTML]{EFEFEF}\textbf{49.13}                                            &
      \cellcolor[HTML]{EFEFEF}\textbf{43.66}                                            &
      \cellcolor[HTML]{EFEFEF}\textbf{40.50}                                            &
      \cellcolor[HTML]{EFEFEF}\textbf{39.83}                                              \\ \bottomrule
    \end{tabular}%
  }
  \vspace{-0.1cm}
\end{table*}

\begin{figure*}[t]
  \centering
  \includegraphics[width=0.9\textwidth]{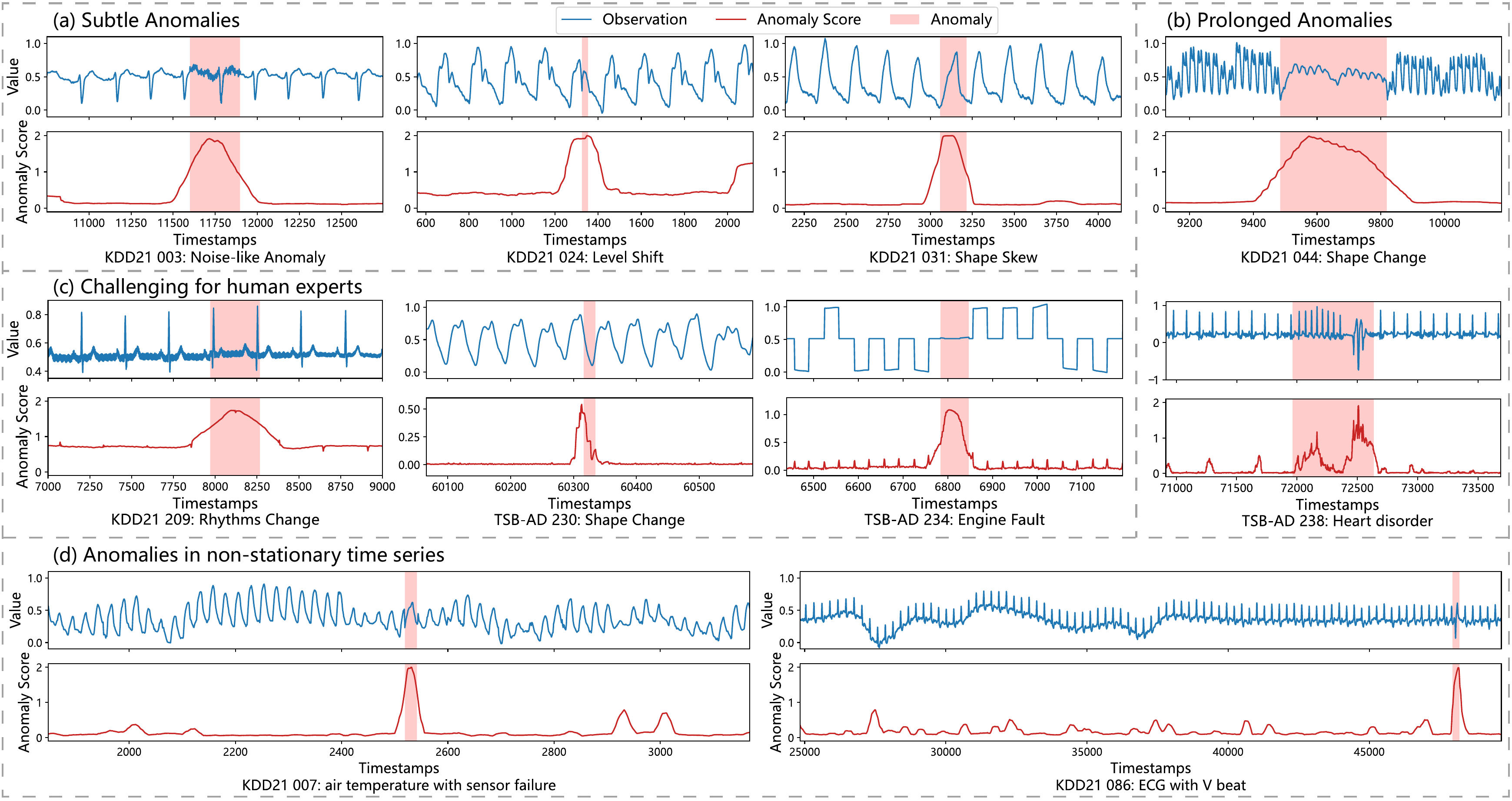}
  \caption{Visualization of detection results of \textsc{CoAD} on challenging cases.}
  \label{fig:detection_example}
  \vspace{-0.5cm}
\end{figure*}

To evaluate the effectiveness of each component and design choice in \textsc{CoAD}, we conduct a comprehensive study with the model variants listed in Table \ref{tab:ablation_study}. Specifically, we examine standalone models (OE or MAE alone), different cooperative strategies, classification granularities, time-frequency ensemble strategies, and anomaly scoring methods. For clarification: \textit{\ul{OE+MAE (Random/Grating)}} directly combines detection results from OE and random/grating masking MAE without guidance;
\textit{\ul{OE+MAE (Guide w/ Hard Mask)}} uses OE guidance with discrete hard masks;
\textit{\ul{OE+MAE (Guide w/o Score)}} guides MAE using OE soft masking, but only uses MAE's reconstruction error for final anomaly scoring;
\textit{\ul{Feature\_Gate}} fuses time and frequency features using a gated mechanism \cite{arevalo2017gatedmultimodalunitsinformation};
and \textit{\ul{Decision\_Mean}} averages the classification scores from both branches. \textbf{Implementation details and descriptions of all variants are available in Appendix \ref{sec:ablation_settings}.}

Based on the results, we draw the following key conclusions:
\textit{i)} \textsc{CoAD} achieves the best overall performance, validating the effectiveness of our design.
\textit{ii)} Cooperative models generally outperform standalone ones, with our guided soft masking framework outperforming hard-masked or naive cooperative baselines.
\textit{iii)} Patch-level classification granularity yields markedly better results than step- or window-level approaches.
\textit{iv)} Frequency-domain information significantly enhances anomaly detection performance.
\textit{vi)} Decision-level fusion via maximum operation is more effective than feature-level fusion or averaging strategies. \textbf{Qualitative ablation study results are available in Appendix \ref{sec:Qualitative}.}

\subsection{Visualization on Challenging Anomalies} \label{sec:Visualization}

Figure~\ref{fig:detection_example} visualizes the detection results of \textsc{CoAD} on several challenging cases:
\textbf{1) Subtle anomalies:} KDD21 003, KDD21 024, and KDD21 031 contain subtle anomalies whose amplitudes are similar to those of normal values, making them easily overlooked~\cite{leeExplainableTimeSeries2024,sunUnravelingAnomalyTime2024}. In contrast, these anomalies can be effectively detected by \textsc{CoAD}.
\textbf{2) Prolonged anomalies:} KDD21 044 and TSB-AD 238 include anomalies that persist over multiple periods, posing difficulties for many existing methods~\cite{mejriUnsupervisedAnomalyDetection2024}. Nevertheless, \textsc{CoAD} successfully captures and localizes these long-duration anomalies.
\textbf{3) Anomalies challenging for human experts:} As emphasized by~\cite{2024Fundamental}, a desirable anomaly detection model should be able to identify anomalies that are difficult even for human experts. KDD21 209, TSB-AD 230, and TSB-AD 234 exemplify such cases. \textsc{CoAD} consistently assigns high anomaly scores to these regions, highlighting its robustness in handling complex and ambiguous patterns.
\textbf{4) Anomalies in non-stationary time series:} As highlighted in \cite{kdd24concept}, existing deep learning-based methods generally fail to detect anomalies in non-stationary series. In contrast, \textsc{CoAD} effectively handles such dynamic cases.

\section{Conclusion}\label{sec:conclusion}
This paper proposes \textsc{CoAD}, a cooperative framework that seamlessly integrates classification and reconstruction to leverage their complementary strengths and overcome their individual limitations. Extensive experiments on reliable datasets using rigorous evaluation metrics validate that our proposed framework significantly outperforms baselines in both detection performance and computational efficiency.

\begin{acks}
    This work was supported by the National Natural Science Foundation of China (72571279) and the science and technology innovation Program of Hunan Province (2023RC1002).
\end{acks}

\bibliographystyle{ACM-Reference-Format}
\balance
\bibliography{ref}

\appendix

\begin{figure*}[h]
  \centering
  \includegraphics[width=0.9\textwidth]{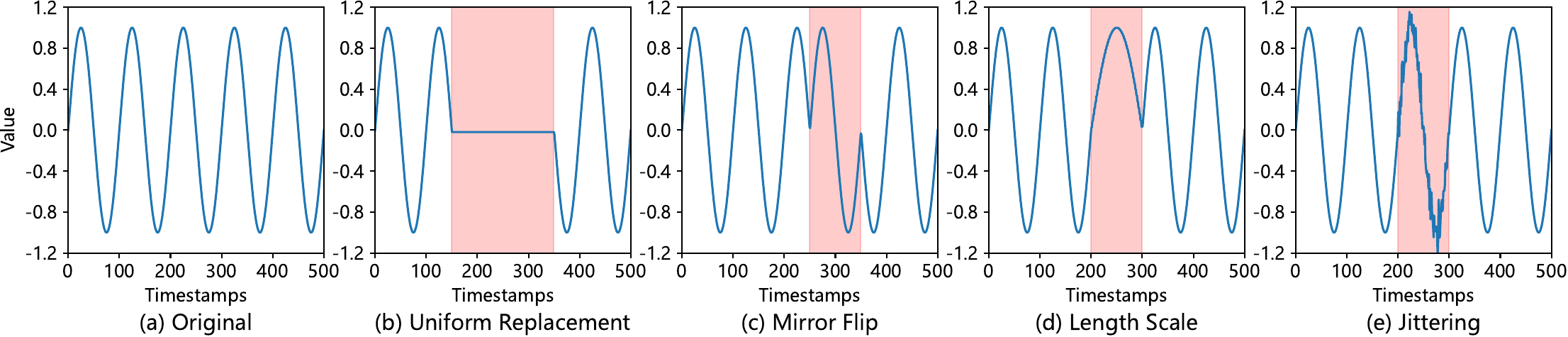}
  \caption{Illustration of the four types of distortions. The red segments represent the distorted parts.}
  \label{fig:simulated_anomalies}
\end{figure*}

\section{Simulated Anomalies}\label{sec:simulated_anomalies}
We simulated anomalies by distorting a portion of the input time series window $\mathbf{X}$. In detail, we randomly select an interval $ [t^{\prime}_1, t^{\prime}_2] \subset [t_1, t_2]$ in the input window $ \mathbf{X} = \{ x_{t_1} \ldots x_{t_2} \} $, and replace the values $ \mathbf{X^{\prime}}_{[t^{\prime}_1, t^{\prime}_2]} = \{ x_{t^{\prime}_1} \ldots x_{t^{\prime}_2} \} $ with one of the following anomalies (see Figure \ref{fig:simulated_anomalies}):
\begin{itemize}[leftmargin=*,topsep=0pt,itemsep=0pt,partopsep=0pt,parsep=0pt,labelindent=\parindent]
  \item \underline{Uniform Replacement:} The original values are replaced with a constant sequence with values in the range $ \{ min(\mathbf{X^{\prime}}), max(\mathbf{X^{\prime}}) \} $.
  \item \underline{Mirror Flip:} The original values are flipped across the x-axis or the y-axis.
  \item \underline{Length Scale:} The original sequences are substituted with lengthened or shortened versions of themselves.
  \item \underline{Jittering:} The original values are added with random noise sampled from a normal distribution $\mathcal{N}(0, 0.1\boldsymbol{I})$.
\end{itemize}

\section{Datasets and Metrics}
\subsection{Datasets}\label{sec:datasets}
The KDD21 benchmark \cite{wuCurrentTimeSeries2021}, also known as the UCR Anomaly Archive, is widely recognized as the highest-quality benchmark for time series anomaly detection ~\cite{tang2025MMMA, sunUnravelingAnomalyTime2024, luMatrixProfileXXIV2022, TimeSeriesBench}. It comprises 250 datasets drawn from diverse domains such as healthcare, sports, industry, and robotics. Notably, the anomalies in the datasets are challenging to detect and could not be easily addressed by the ``one-liner'' approach \cite{wuCurrentTimeSeries2021}. In addition, the KDD21 benchmark provides a document explaining why certain regions are labeled as anomalies. This further enhances the credibility and transparency of the dataset.

The TSB-AD benchmark \cite{liu2024the} represents the largest currently available collection of time-series anomaly detection datasets. It curates and manually cleanses datasets from various sources. However, it has a significant overlap with the KDD21 dataset, and several subsets, such as NAB, WSD, YAHOO, and Stock, still suffer from issues including mislabeled ground truth, trivial anomalies, and unrealistic anomaly densities \cite{2024Fundamental}. Therefore, we exclude the overlapping parts and adhere to the criteria outlined in prior research on dataset quality \cite{wuCurrentTimeSeries2021, luDAMPAccurateTime2023,schmidlAnomalyDetectionTime2022} to select several high-quality subsets from the TSB-AD dataset. The selected subsets include MGAB, SED, SVDB, IOPS, and TODS.

The statistics of the datasets are summarized in Table \ref{tab:dataset_statistics}. The anomaly ratio is calculated from the ratio between the sum of all anomaly points and the sum of all test points.

\begin{table}[htbp]
  \centering
  \caption{Statistics of benchmarks.}
  \begin{tabularx}{0.9\columnwidth}{@{} l *{2}{>{\centering\arraybackslash}X} @{}}
    \toprule
    \textbf{Benchmark}     & {\centering \textbf{KDD21}} & {\centering \textbf{TSB-AD}} \\
    \midrule
    Num of Datasets        & \num{250}                   & \num{64}                     \\
    Overall Train Length   & \num{2238250}               & \num{778432}                 \\
    Overall Test Length    & \num{6143541}               & \num{5690048}                \\
    Average Anomaly Length & \num{147}                   & \num{56}                     \\
    Anomaly Ratio          & 0.60\%                      & 2.70\%                       \\
    \bottomrule
  \end{tabularx}
  \label{tab:dataset_statistics}
\end{table}

\subsection{Metrics}\label{sec:metrics}
Recent studies \cite{liu2024the, tang2025MMMA, kimRigorousEvaluationTimeSeries2022} indicate that widely used evaluation metrics, such as point-adjusted F1 (F1-PA) and F1-Affiliation, tend to significantly overestimate model performance, even assigning high evaluation scores to randomly generated predictions. In addition, anomaly detection datasets are highly class-imbalanced, with normal points far outnumbering anomalous points. The AUC-ROC (Area Under the Receiver Operating Characteristic Curve) metric is biased towards the majority class, leading to inflated evaluation scores \cite{davisRelationshipPrecisionRecallROC2006}. AUC-PR (Area Under the Precision-Recall Curve) has been advocated as a more informative alternative for imbalanced datasets \cite{2010AUC}. Therefore, a recent benchmark evaluation paper has demonstrated that Standard-F1, AUC-PR \cite{davisRelationshipPrecisionRecallROC2006}, Range-AUC-PR \cite{paparrizosVolumeSurfaceNew2022} and VUS-PR \cite{paparrizosVolumeSurfaceNew2022} are the most reliable and accurate metrics for assessing model performance. The details of the evaluation metrics are described as follows:
\begin{itemize}[leftmargin=*,topsep=0pt,itemsep=0pt,partopsep=0pt,parsep=0pt,labelindent=\parindent]
  \item \ul{Standard-F1} is computed directly from the obtained anomaly scores without any post-processing. To reduce the influence of threshold selection on evaluation results and maintain consistency with prior works \cite{liu2024the,gargEvaluationAnomalyDetection2022,kimRigorousEvaluationTimeSeries2022,liEmpiricalAnalysisAnomaly2023}, we report the maximum F1-score across all possible thresholds.
  \item \ul{AUC-PR}~\cite{davisRelationshipPrecisionRecallROC2006} measures the area under the precision-recall curve, providing a threshold-independent evaluation of model performance.
  \item \ul{Range-AUC-PR}~\cite{paparrizosVolumeSurfaceNew2022} addresses labeling uncertainty and anomaly scoring delays by introducing a buffer zone around the boundaries of labeled anomalies when calculating AUC-PR, thereby giving some credit to high anomaly scores in the vicinity of the anomaly boundaries. Following the original study, we set the buffer length to the average anomaly length for each dataset.
  \item \ul{VUS-PR}~\cite{paparrizosVolumeSurfaceNew2022} further solves the buffer length selection issue in Range-AUC-PR by calculating the volume under the surface formed by varying buffer lengths and thresholds, thus providing a more robust evaluation metric.
\end{itemize}

\section{Implementation Details}\label{sec:implementation}
Both our model and all baselines follow identical data preprocessing procedures within an integrated, unified pipeline. All experiments are conducted on an Intel i9-12900K CPU and a single NVIDIA RTX 3090 GPU.

\subsection{Implementation Details of \textsc{CoAD}}
\textsc{CoAD} takes the Gate Recurrent Unit (GRU) \cite{2014Empirical} as the encoder, with a hidden size of 24 and 3 recurrent layers. The input window size $\mathtt{T}$ is set to 4 times the dominant period of the time series, while the patch size $\mathtt{P}$ is fixed to 8. The dominant period is automatically determined using the autocorrelation function (ACF). The loss weight $\lambda$ is set to 10, and the number of frequency bands $\mathtt{K}$ for STFT is set to 4. We use the Adam optimizer with a learning rate of 0.002. We train our model for 300 epochs for all datasets.
\textbf{A detailed analysis of the hyperparameters is presented in Section~\ref{sec:parameter}.}

\subsection{Implementation Details of Baselines}
\ul{\textit{For deep learning-based methods},} we reproduce all models using their official open-source repositories. We strictly follow the original training and test splits provided by the KDD21 and TSB-AD benchmarks and employ early stopping for model selection. Each deep learning-based method is trained 5 times with different random seeds, and the averaged performance is reported.

\ul{\textit{For data mining-based methods},} we include the 7 best-performing approaches identified in recent evaluation studies~\cite{liu2024the}. All implementations are based on the TSB-UAD \cite{paparrizosTSBUADEndtoEndBenchmark2022} and TSB-AD \cite{liu2024the} libraries. The hyperparameters for these methods are set according to the TSB-AD \cite{liu2024the} paper, where they are tuned on a large-scale validation set.

\begin{table}[t]
  \caption{The evaluation results (\%) on the KDD21 dataset using Top-$\mathit{k}$ Accuracies (Acc.@$\mathit{k}$). The best results are in \textbf{bold}, and the second-best results are with \underline{underline}.}
  \label{tab:KDD21_score}
  \renewcommand{\arraystretch}{0.85}
  \resizebox{\columnwidth}{!}{%
    \begin{tabular}{@{}cc|ccc@{}}
      \toprule
      \multicolumn{2}{c|}{\textbf{Model / Dataset}}                                                                  & \multicolumn{3}{c}{\textbf{KDD21}}                                        \\ \midrule
      \multicolumn{1}{c|}{Model Class}                                                                               & Model                              & Acc.@1     & Acc.@3     & Acc.@5     \\ \midrule
      \multicolumn{1}{c|}{}                                                                                          & DADA                               & 7.47       & 11.62      & 17.84      \\
      \multicolumn{1}{c|}{}                                                                                          & MOMENT                             & 15.70      & 21.49      & 25.21      \\
      \multicolumn{1}{c|}{}                                                                                          & MMA                                & 40.00      & 52.40      & 58.40      \\
      \multicolumn{1}{c|}{\multirow{-4}{*}{Pure MAE}}                                                                & TFMAE                              & 3.48       & 7.39       & 12.61      \\ \midrule
      \multicolumn{1}{c|}{}                                                                                          & AnomalyBERT                        & 9.64       & 15.66      & 57.43      \\
      \multicolumn{1}{c|}{}                                                                                          & CutAddPaste                        & 55.20      & 64.80      & 71.20      \\
      \multicolumn{1}{c|}{}                                                                                          & TriAD                              & 25.60      & 34.00      & 34.00      \\
      \multicolumn{1}{c|}{\multirow{-4}{*}{Pure OE}}                                                                 & COUTA                              & 15.48      & 24.27      & 28.45      \\ \midrule
      \multicolumn{1}{c|}{}                                                                                          & FCVAE                              & 37.60      & 48.40      & 56.00      \\
      \multicolumn{1}{c|}{}                                                                                          & TFAD                               & 4.13       & 6.61       & 11.57      \\
      \multicolumn{1}{c|}{\multirow{-3}{*}{\begin{tabular}[c]{@{}c@{}}Time-Frequency\\ Reconstruction\end{tabular}}} &
      CATCH                                                                                                          &
      33.06                                                                                                          &
      45.04                                                                                                          &
      50.41                                                                                                                                                                                      \\ \midrule
      \multicolumn{1}{c|}{}                                                                                          & TranAD                             & 24.00      & 29.60      & 33.60      \\
      \multicolumn{1}{c|}{}                                                                                          & MAUT                               & 30.40      & 39.20      & 43.60      \\
      \multicolumn{1}{c|}{}                                                                                          & M2N2                               & 11.98      & 17.36      & 21.49      \\
      \multicolumn{1}{c|}{}                                                                                          & FITS                               & 28.92      & 36.36      & 43.38      \\
      \multicolumn{1}{c|}{}                                                                                          & DCdetector                         & 16.13      & 27.42      & 32.26      \\
      \multicolumn{1}{c|}{\multirow{-6}{*}{Other Deep Learning}}                                                     &
      AnomalyTrans                                                                                                   &
      8.47                                                                                                           &
      17.34                                                                                                          &
      21.37                                                                                                                                                                                      \\ \midrule
      \multicolumn{1}{c|}{}                                                                                          & KshapeAD                           & 46.00      & 58.80      & 65.60      \\
      \multicolumn{1}{c|}{}                                                                                          & SAND                               & 47.20      & 58.40      & 64.80      \\
      \multicolumn{1}{c|}{}                                                                                          & Sub-PCA                            & 21.68      & 28.92      & 34.14      \\
      \multicolumn{1}{c|}{}                                                                                          & Series2Graph                       & 36.40      & 50.00      & 55.20      \\
      \multicolumn{1}{c|}{}                                                                                          & KmeansAD                           & 40.00      & 52.80      & 59.60      \\
      \multicolumn{1}{c|}{}                                                                                          & Matrix Profile                     & \ul{56.80} & \ul{72.40} & \ul{78.40} \\
      \multicolumn{1}{c|}{\multirow{-7}{*}{Data Mining}}                                                             & DAMP                               & 38.40      & 44.40      & 51.20      \\ \midrule
      \rowcolor[HTML]{EFEFEF}
      \multicolumn{1}{c|}{\cellcolor[HTML]{EFEFEF}\textbf{Cooperative}}                                              &
      \textbf{\textsc{CoAD} (ours)}                                                                                  &
      \textbf{61.60}                                                                                                 &
      \textbf{76.80}                                                                                                 &
      \textbf{80.80}                                                                                                                                                                             \\ \bottomrule
    \end{tabular}%
  }
  \vspace{-0.3cm}
\end{table}

\section{Evaluation Results on the KDD21 Benchmark Using Top-$\mathit{k}$ Accuracy Scores} \label{sec:KDD21_score}
\underline{\textit{Settings.}} The KDD21 benchmark originates from the SIGKDD Cup 2021 competition, and we also adopt the official evaluation metric provided by the competition organizer \cite{kdd21competition}.
Each of the 250 datasets in the KDD21 benchmark contains only a single anomaly. For each algorithm, the point with the highest anomaly score (Top-$\mathit{1}$) is selected as the predicted anomaly. A prediction is considered correct if it lies within $\pm100$ data points of the true anomaly range; otherwise, it is marked incorrect.
The overall accuracy is computed as the mean accuracy across all datasets. In addition to Top-$\mathit{1}$ accuracy, we also report Top-$\mathit{3}$ and Top-$\mathit{5}$ accuracies to account for scenarios where multiple plausible anomaly locations may exist in real-world time series.

\underline{\textit{Results.}} Table~\ref{tab:KDD21_score} summarizes the evaluation results on the KDD21 benchmark in terms of Top-$\mathit{k}$ accuracy (Acc.@$\mathit{k}$). The empirical data consistently demonstrate the superior performance of the proposed \textsc{CoAD} framework.

\section{Efficiency Experiments on the TSB-AD Benchmark} \label{sec:tsb_ad_efficiency}
\subsection{Experiment Settings} We comprehensively compare the detection performance, inference speed, and model params of \textsc{CoAD} against baseline methods. Following existing works \cite{tang2025MMMA, luMatrixProfileXXIV2022}, we report the total inference time across all subsets as the overall inference time. To ensure a fair comparison, all deep learning-based models are evaluated with the same batch size of 128, the window size that yields the highest VUS-PR score, and are executed on the same NVIDIA RTX 3090 GPU. All Data Mining-based methods are tested on the same Intel i9-12900K CPU, as they don't support GPU parallel processing.

\begin{figure}[t]
  \centering
  \includegraphics[width=\columnwidth]{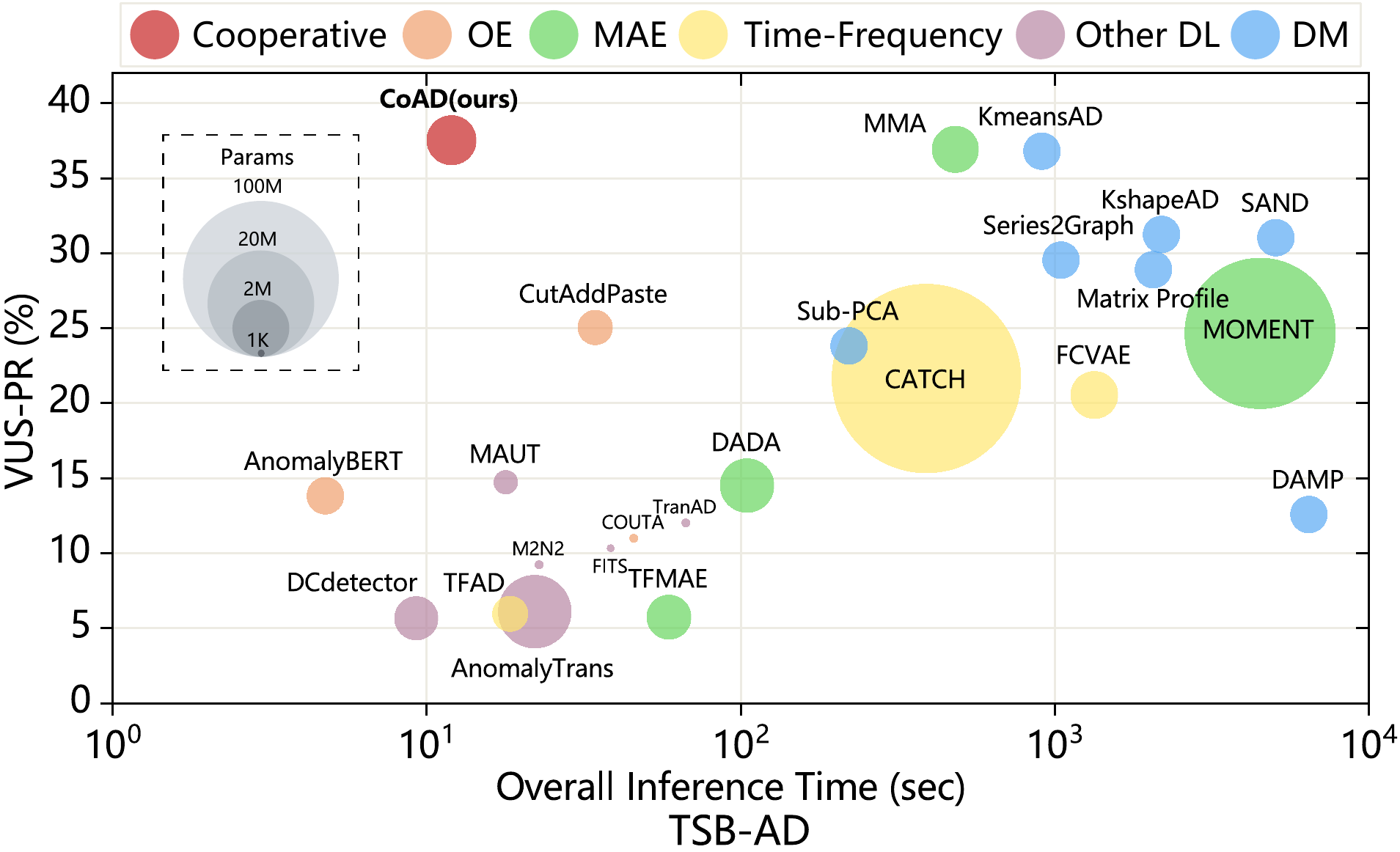}
  \caption{Model efficiency comparison.}
  \label{fig:tsb_efficiency}
  \vspace{-0.5cm}
\end{figure}

\begin{figure*}[t]
  \centering
  \includegraphics[width=\textwidth]{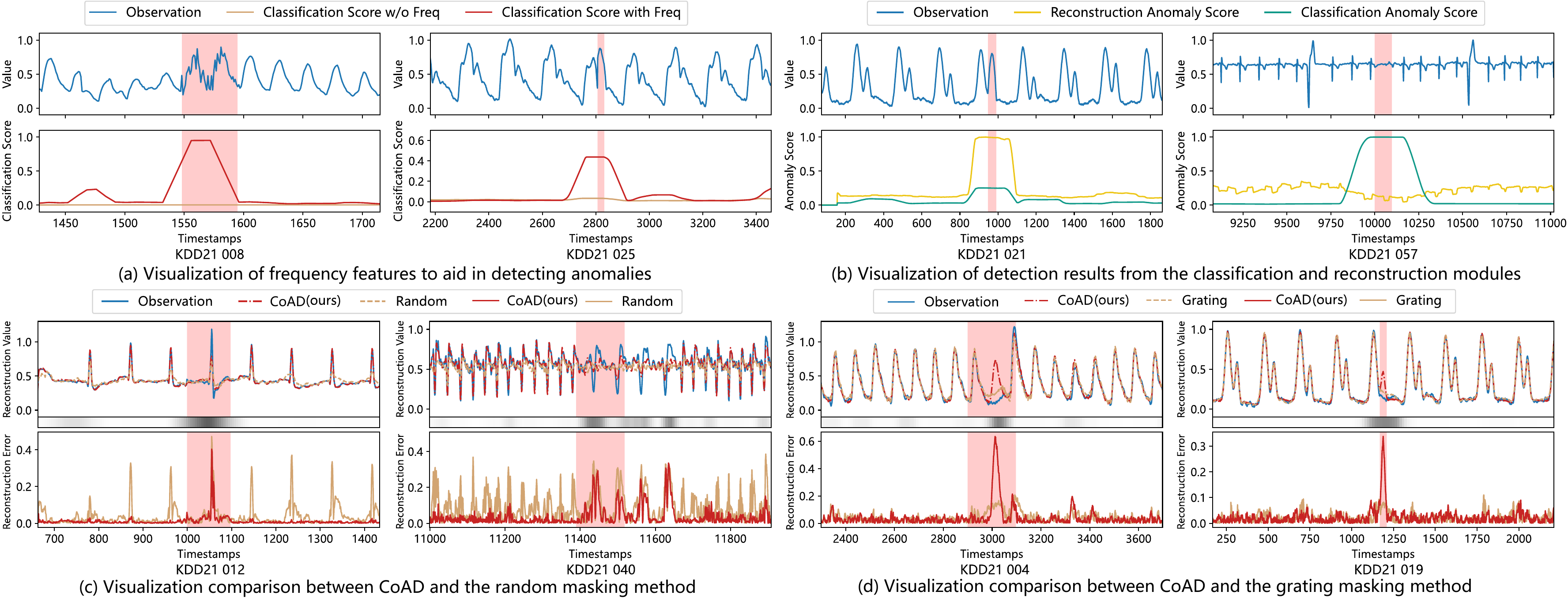}
  \caption{Qualitative ablation results. The lower parts of (c) and (d) represent the probability-informed soft masks.}
  \label{fig:submodule_compare}
\end{figure*}

\subsection{Efficiency Results} The model efficiency comparison on the TSB-AD dataset is shown in Figure \ref{fig:tsb_efficiency}. Our proposed framework \textsc{CoAD} outperforms the state-of-the-art methods in terms of both efficiency and performance. While MMA and KmeansAD demonstrate performance levels comparable to \textsc{CoAD}, \textsc{CoAD} achieves inference speeds that are several orders of magnitude faster.
Specifically, \textsc{CoAD} completes the inference on all subsets of the TSB-AD benchmark, comprising more than 5.69 million data points, in just 37.55 seconds.

\section{Time Complexity Analysis} \label{sec:time_complex}
The computational complexity of \textsc{CoAD} primarily arises from its classification and reconstruction modules. In the classification module, the frequency branch involves STFT computation with time complexity $O(\mathtt{T})$. The GRU encoders in both the frequency and time branches have a time complexity of $O( \frac{\mathtt{T}}{\mathtt{P}} \cdot \mathtt{H}^2)$. The reconstruction module's GRU encoder also has a time complexity of $O( \frac{\mathtt{T}}{\mathtt{P}} \cdot \mathtt{H}^2)$. Therefore, the overall time complexity of \textsc{CoAD} is $O\left(\mathtt{T} + \frac{\mathtt{T}}{\mathtt{P}} \cdot \mathtt{H}^2\right)$. This indicates that the computational cost scales linearly with the input sequence length $\mathtt{T}$, while the patching design effectively reduces $\mathtt{T}$ by a factor of the patch length $\mathtt{P}$, thereby substantially reducing computational overhead.

\section{Detailed Ablation Settings} \label{sec:ablation_settings}
The implementation details of ablation variants are outlined below:
\begin{enumerate}[leftmargin=*,label=\textit{\roman*)}]
  \item \textit{\ul{OE and MAE variants works alone:}}
        \begin{itemize}[leftmargin=*]
          \item \textit{OE (Time)} takes only the time domain features for classification.
          \item \textit{OE (Time + Frequency)} employs the maximum fusion strategy to combine classification results from both the time and frequency domains.
          \item \textit{MAE (Random/Grating)} applies either the random or grating masking strategy.
        \end{itemize}
  \item \textit{\ul{Different cooperative strategies:}}
        \begin{itemize}[leftmargin=*]
          \item \textit{OE+MAE (Random/Grating)} calculates the anomaly score by summing the anomaly probability produced by OE and the reconstruction error generated by \textit{MAE (Random/Grating)}.
          \item \textit{OE+MAE (Guide w/ Hard Mask)} integrates OE's guidance using discrete hard masks, where patches with anomaly probabilities exceeding a certain threshold are masked. The threshold is determined as the mean plus three standard deviations of the anomaly probabilities across all patches in the training set.
          \item \textit{OE (Time) + MAE (Guide w/ Soft Mask)} is softly guided by the OE module; however, the OE module performs classification based solely on time-domain features.
        \end{itemize}
  \item \textit{\ul{Cooperation under different design choices:}}
        \begin{itemize}[leftmargin=*]
          \item \textit{OE (Step Level)+MAE} adopts step-level classification granularity similar to AnomalyBERT \cite{jeongAnomalyBERTSelfSupervisedTransformer2023}. The mean anomaly probability across all points within a patch is used as the patch's anomaly probability for soft masking.
          \item \textit{OE (Window Level)+MAE} utilizes window-level classification granularity similar to CutAddPaste \cite{wangCutAddPasteTimeSeries2024}. A sliding window with a stride of 1 is applied to obtain the anomaly probability for each point. The soft masking process is then performed as in \textit{OE (Step Level)+MAE}.
          \item \textit{OE (Feature Add)+MAE} employs a feature-level fusion strategy, where features from both domains are directly added and then fed into the classifier.
          \item \textit{OE (Feature Gated)+MAE} also uses a feature-level fusion strategy, combining features from both domains through a learnable gating mechanism \cite{arevalo2017gatedmultimodalunitsinformation}.
          \item \textit{OE (Decision Mean)+MAE} applies a decision-level fusion strategy, averaging the classification scores from both branches.
          \item \textit{OE+MAE (Guide w/o Score)} takes the same guided soft masking strategy as \textsc{CoAD}, but relies solely on MAE's reconstruction error for final anomaly scoring.
        \end{itemize}
\end{enumerate}

\begin{figure*}[!t]
  \centering
  \includegraphics[width=\textwidth]{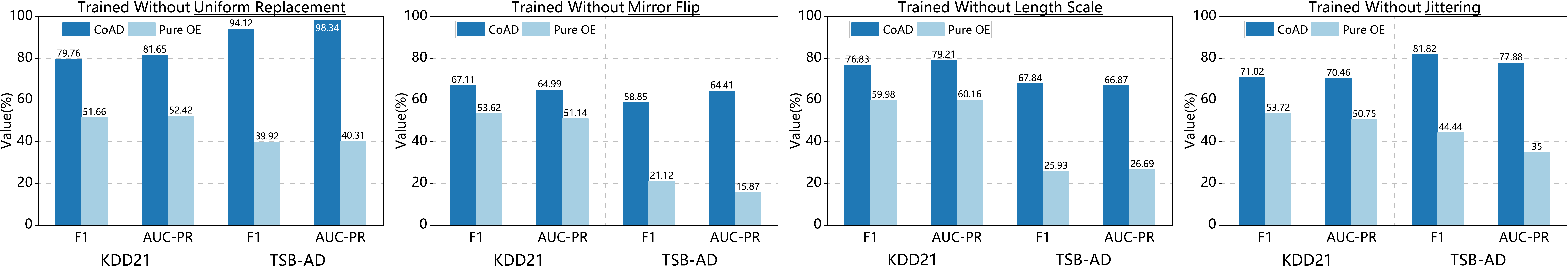}
  \caption{Comparison of \textsc{CoAD} and Pure OE in detecting unseen anomalies. The experiments are conducted under cross-type settings (e.g., trained without the \textit{Uniform Replacement} distortion type and tested on \textit{Uniform Replacement} anomalies).}
  \label{fig:generalization_study}
  \vspace{-0.3cm}
\end{figure*}

\begin{figure*}[t]
  \centering
  \subfloat[Parameter analysis on the KDD21 benchmark with respect to the window size, patch size, loss weight~$\lambda$ and backbone choice.]{
    \includegraphics[width=\textwidth]{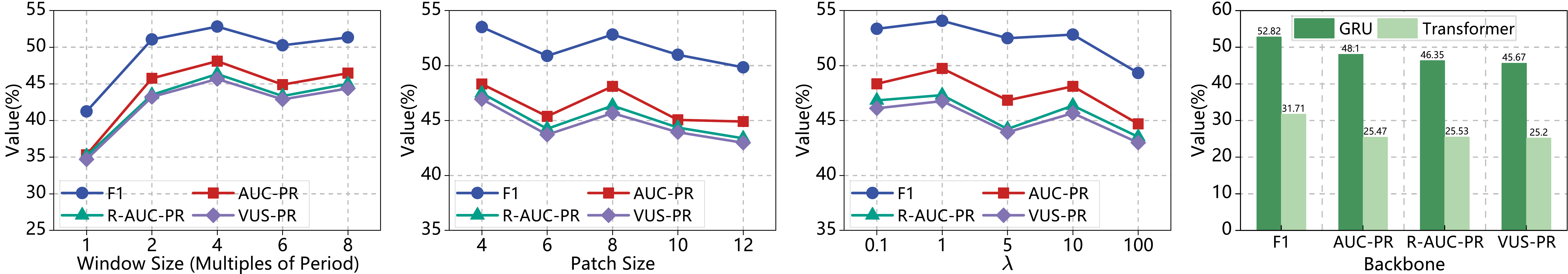}
    \vspace{-0.5cm}
    \label{fig:KDD21_param}
  }
  \vspace{-0.3cm}
  \subfloat[Parameter analysis on the TSB-AD benchmark with respect to the window size, patch size, loss weight~$\lambda$ and backbone choice.]{
    \includegraphics[width=\textwidth]{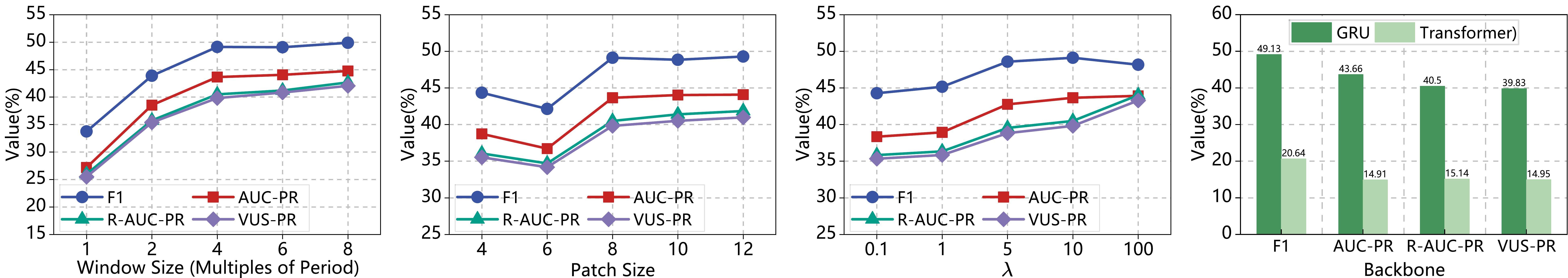}
    \vspace{-0.5cm}
    \label{fig:TSB_AD_param}
  }
  \caption{Hyperparameter study on (a) KDD21 and (b) TSB-AD datasets.}
  \label{fig:parameter_study}
\end{figure*}

\section{Qualitative Ablation Results} \label{sec:Qualitative}
We further present qualitative ablation results in Figure~\ref{fig:submodule_compare}, providing an intuitive understanding of the contribution and effectiveness of each component in \textsc{CoAD}.
Figure~\ref{fig:submodule_compare}(a) clearly shows that incorporating frequency-branch classification results helps identify anomalies that are difficult to detect in the time domain alone.
Figure~\ref{fig:submodule_compare}(b) confirms that joint anomaly inference from both the classification and reconstruction modules delivers more comprehensive detection results than relying solely on either module individually.
Figure~\ref{fig:submodule_compare}(c) and (d) highlight the superiority of our proposed guided soft masking strategy over existing random and grating masking methods.
The random masking approach generates high reconstruction errors even in normal regions, leading to severe false alarms, while the grating masking method tends to overfit anomalies, resulting in false negatives.
Our proposed probability-informed soft masking suppresses anomaly-related cues and retains normal information, enabling accurate reconstruction in normal regions and higher reconstruction errors in anomalous regions, thereby maximizing the reconstruction module’s performance.

\section{Generalizability Verification} \label{sec:generalization}
We conduct experiments to evaluate the generalizability of \textsc{CoAD} in comparison with the \textit{pure OE-based method}. \textsc{CoAD} incorporates anomaly scores computed from both the Time-Frequency ensemble classification and the Residual classification modules, whereas the \textit{pure OE-based method} computes anomaly scores solely from the Time–Frequency ensemble classification module. Specifically, one of the four simulated anomaly types in Figure \ref{fig:simulated_anomalies} is excluded during training and then reintroduced in the testing set to replace the original anomalies, enabling the assessment of each model’s ability to detect entirely novel anomaly types. We report results using the Standard-F1 and AUC-PR metrics, as the remaining metrics exhibit consistent trends.

The evaluation results are presented in Figure~\ref{fig:generalization_study}. The \textit{pure OE-based method} shows unstable performance when encountering unseen anomaly types, with particularly poor results on \textit{Mirror Flip} and \textit{Length Scale} in the TSB-AD benchmark. In contrast, \textsc{CoAD} demonstrates consistently robust performance across all test scenarios, indicating that its cooperative design effectively enhances generalization and enables reliable detection of previously unseen anomaly types.

\section{Hyperparameter Study}
\label{sec:parameter}
We conduct a comprehensive parameter study to investigate the sensitivity of \textsc{CoAD} to key hyperparameters, including the input window size $\mathtt{T}$, patch size $\mathtt{P}$, and loss weight $\lambda$. The results are presented in Figure~\ref{fig:parameter_study}. The following observations can be made: 1) Increasing the input window size generally improves model performance. This is because a larger window enables the model to leverage richer contextual information to facilitate anomaly detection. However, when the window size exceeds 4 times the dominant period, the performance improvement becomes marginal while introducing additional computational overhead. 2) The performance of \textsc{CoAD} initially improves with increasing patch size, as larger patches provide more comprehensive local context for both classification and reconstruction. However, excessively large patches may lead to over-smoothing and the loss of fine-grained details \cite{tangUnlockingPowerPatch}, leading to a slight decline in performance. 3) Since the classification loss and reconstruction loss are on different scales, the loss weight $\lambda$ is critical to balance the learning process. In practice, setting $\lambda$ between 5 and 10 generally yields optimal performance.

We further investigate different backbone choices with results shown on the right side of Figure~\ref{fig:parameter_study}. \textsc{CoAD} with GRU \cite{2014Empirical} as the backbone achieves significantly better performance than Transformer \cite{vaswaniAttentionAllYou2017}. This is consistent with prior findings \cite{Zeng_Chen_Zhang_Xu_2023, Tang_Zhang_2025} that the self-attention mechanism inevitably leads to a loss of temporal information in time series data (i.e., permutation-invariant and anti-order characteristics of self-attention), thereby impairing anomaly detection performance. Meanwhile, recent studies also demonstrate that the~\textbf{Patch + RNN} architecture is more effective in time series modeling~\cite{Kong_Wang_Nie_Zhou_Zohren_Liang_Sun_Wen_2025,lin2023segrnnsegmentrecurrentneural}, as the inherent sequential structure of RNN models enables them to effectively capture the crucial temporal dependencies, and the patching design significantly improves the long-term temporal modeling ability and computational efficiency for RNNs.

\end{document}